\documentclass[acmsmall]{acmart}
\usepackage{booktabs,multirow,array}
\usepackage{bm}
\AtBeginDocument{%
  \providecommand\BibTeX{{%
    \normalfont B\kern-0.5em{\scshape i\kern-0.25em b}\kern-0.8em\TeX}}}

\setcopyright{acmcopyright}
\acmJournal{TOIS}
\acmYear{2021} \acmVolume{1} \acmNumber{1} \acmArticle{1} \acmMonth{1} \acmPrice{15.00}\acmDOI{10.1145/3507782}

\acmJournal{TOIS}
\acmVolume{40}
\acmNumber{2}
\acmArticle{111}
\acmMonth{4}

\begin{document}

\title{Towards Personalized Answer Generation in E-Commerce via Multi-Perspective Preference Modeling}\thanks{The work described in this article is substantially supported supported by a grant from the Research Grant Council of the Hong Kong Special Administrative Region, China (Project Code: 14200719), and Alibaba Group through Alibaba Research Intern Program. Work done when Yang Deng was an intern at Alibaba.}

\author{Yang Deng}
\email{ydeng@se.cuhk.edu.hk}
\affiliation{%
  \institution{The Chinese University of Hong Kong}
  \city{Hong Kong}
  \postcode{999077}
}

\author{Yaliang Li}
\email{yaliang.li@alibaba-inc.com}
\affiliation{%
  \institution{Alibaba Group}
  \city{Bellevue, WA}
  \country{USA}}

\author{Wenxuan Zhang}
 \email{wxzhang@se.cuhk.edu.hk}
\affiliation{%
 \institution{The Chinese University of Hong Kong}
 \city{Hong Kong}}

\author{Bolin Ding}
\email{bolin.ding@alibaba-inc.com}
\affiliation{%
  \institution{Alibaba Group}
  \city{Bellevue, WA}
  \country{USA}}

\author{Wai Lam}
 \email{wlam@se.cuhk.edu.hk}
\affiliation{%
 \institution{The Chinese University of Hong Kong}
 \city{Hong Kong}}

\renewcommand{\shortauthors}{Deng, et al.}

\begin{abstract}
Recently, Product Question Answering (PQA) on E-Commerce platforms has attracted increasing attention as it can act as an intelligent online shopping assistant and improve the customer shopping experience. 
Its key function, automatic answer generation for product-related questions, has been studied by aiming to generate content-preserving while question-related answers.
However, an important characteristic of PQA, i.e., personalization, is neglected by existing methods.
It is insufficient to provide the same ``completely summarized" answer to all customers, since many customers are more willing to see personalized answers with customized information only for themselves, by taking into consideration their own preferences towards product aspects or information needs. 
To tackle this challenge, we propose a novel \textbf{P}ersonalized \textbf{A}nswer \textbf{GE}neration method (\textbf{PAGE}) with multi-perspective preference modeling, which explores historical user-generated contents to model user preference for generating personalized answers in PQA. 
Specifically, we first retrieve question-related user history as external knowledge to model knowledge-level user preference. Then we leverage Gaussian Softmax distribution model to capture latent aspect-level user preference. Finally, we develop a persona-aware pointer network to generate personalized answers in terms of both content and style by utilizing personal user preference and dynamic user vocabulary. 
Experimental results on real-world E-Commerce QA datasets demonstrate that the proposed method outperforms existing methods by generating informative and customized answers, and show that answer generation in E-Commerce can benefit from personalization. 

\end{abstract}


\maketitle

\section{Introduction}\label{sec:intro}
Product Question Answering (PQA) on E-Commerce platforms, such as Amazon\footnote{\url{https://www.amazon.com}}, eBay\footnote{\url{https://www.ebay.com}}, and Alibaba\footnote{\url{https://www.taobao.com}}, enables online customers to seek useful information when making purchase decisions. However, answering product-related questions relies heavily on the willingness of the customers who have already bought the same product or the tedious workload of online sales representatives. 
In order to address this issue, recent studies~\cite{www16-amazon-qa,icdm16-amazon} aim at building an automatic PQA system that takes a product-related question as input, and automatically provides an answer derived from existing E-Commerce data, such as product reviews, descriptions, attributes, etc.   

Early works in PQA focus on solving the subjective questions~\cite{www16-amazon-qa,icdm16-amazon,wsdm18-answer-pred}, which aim to provide yes/no answers by identifying customers' opinions from relevant reviews towards the given question. Following the general community question answering (CQA) problem setting, some recent studies~\cite{aaai19-answer-sel,kdd19-answer-sel} regard relevant reviews as candidate answers and rank a set of review snippets by measuring the semantic relevance between the question and review snippets. However, the review snippets selected by these methods fall short of answering questions effectively, since the user-posted reviews are the general opinions towards the product but not written specifically for the questions. 
Inspired by the recent success on generative open domain question answering~\cite{acl19-answer-gen,emnlp19-answer-gen}, text generation methods~\cite{wsdm19-answer-gen-chen,wsdm19-answer-gen-gao,ijcai19-amazonqa} have been adopted to generate natural answers to product-related questions. The answers are typically generated by summarizing the supporting evidence from relevant reviews.

\begin{figure*}
\centering
\includegraphics[width=0.6\textwidth]{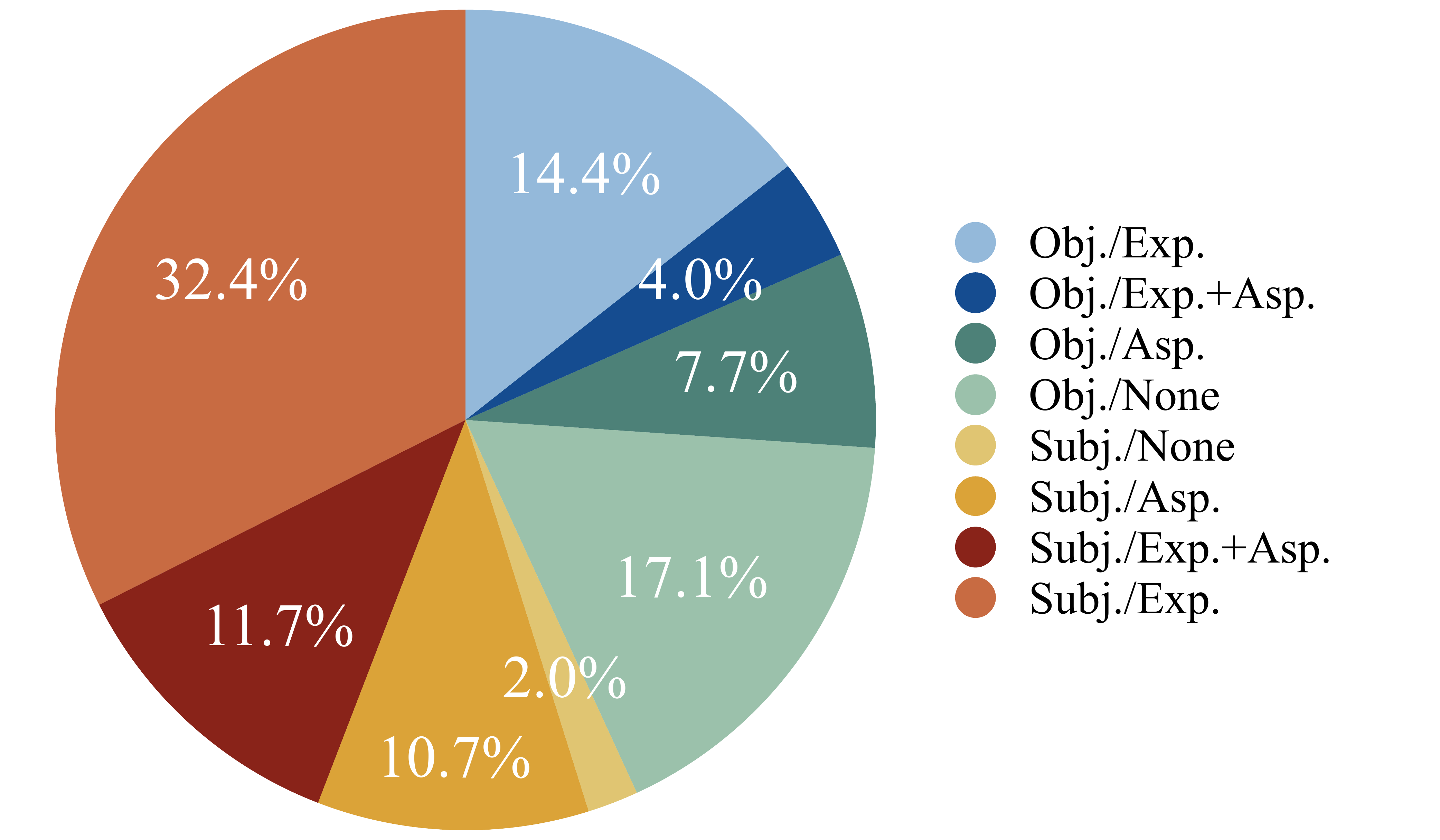}
\caption{The Statistics of User-generated Questions and Answers on Amazon QA (best viewed in color). Subj. and Obj. denote subjective and objective questions respectively. Exp. and Asp. denote the answers with personal experience or preferred product aspects, respectively, while None denotes the answers that cannot reflect any user preference. }
\label{user_study}
\end{figure*}

Despite the effectiveness of these methods on automatically providing answers to product-related questions, they neglect an important characteristic of PQA, i.e., personalization. 
Compared with open-domain question answering~\cite{squad,coling14-cqa}, on PQA platforms, there is a large proportion of subjective questions~\cite{www16-amazon-qa} that involve user preference or require personal information to answer, rather than objective or factoid questions that look for certain unified answer.
Therefore, \citet{sigir18-pqa-challenge} state that a good PQA system should answer the customer’s questions with the context of the customer’s encounter history, taking into consideration his/her preference and interest. 
Such personalization can make the answer more helpful for customers and better clarify their concerns about the product. In order to verify the importance of personalization in PQA, we conduct a statistical user study by analyzing the Amazon QA datasets~\cite{icdm16-amazon}, collected from the real-world PQA platform. The user study is conducted as follows:
\begin{itemize}
    \item[(1)] We randomly sample 100 helpful QA pairs from each of three different product categories, including \textit{Electronics}, \textit{Home\&Kitchen}, \textit{Sports\&Outdoors}, i.e., 300 questions in total.  Note that a helpful QA pair means that the user-provided answer receives more than two votes as well as more upvotes than downvotes. 
    \item[(2)] We manually label the QA pairs by two criteria: (i) Is the question subjective or objective; (ii) Whether the answer reflects any specific user preference? If yes, does the answer reflects the personal preference of user experience or product aspects? 
    \item[(3)]	Each QA pair is labeled by two annotators and the disagreements (about 4\%) are settled by discussions. 
\end{itemize}
The statistics of the user study are presented in Figure~\ref{user_study}. There are several notable observations: \begin{itemize}
    \item[(1)]	There are a larger proportion of subjective questions (56.8\%) in PQA than objective questions (43.2\%), which is similar to the findings in previous studies \cite{www16-amazon-qa,icdm16-amazon}. 
    \item[(2)]	Among these helpful QA pairs, we observe that 80.9\% of answers can somewhat reflect specific personalized information, including personal experience or preferred product aspects. Especially for those subjective questions, almost all the helpful answers involve such personalized information. Moreover, for those objective questions, there are still half of the helpful answers that contain personalized information. 
    \item[(3)]	Among those answers that reflect user preferences, there are two different perspectives of user preferences: one is concerned with personal experience and the other one is concerned with preferred product aspects. Besides, it can be observed that there are also a large number of answers that contain both of them.  
\end{itemize}
To this end, we study personalized answer generation for PQA, which can not only generate natural language answers based on relevant reviews, but also provide customized answers to different customers by exploring and modeling user preference. 
To the best of our knowledge, this work is the first attempt to marry product question answering with personalization.

\begin{figure}
\centering
\includegraphics[width=0.8\textwidth]{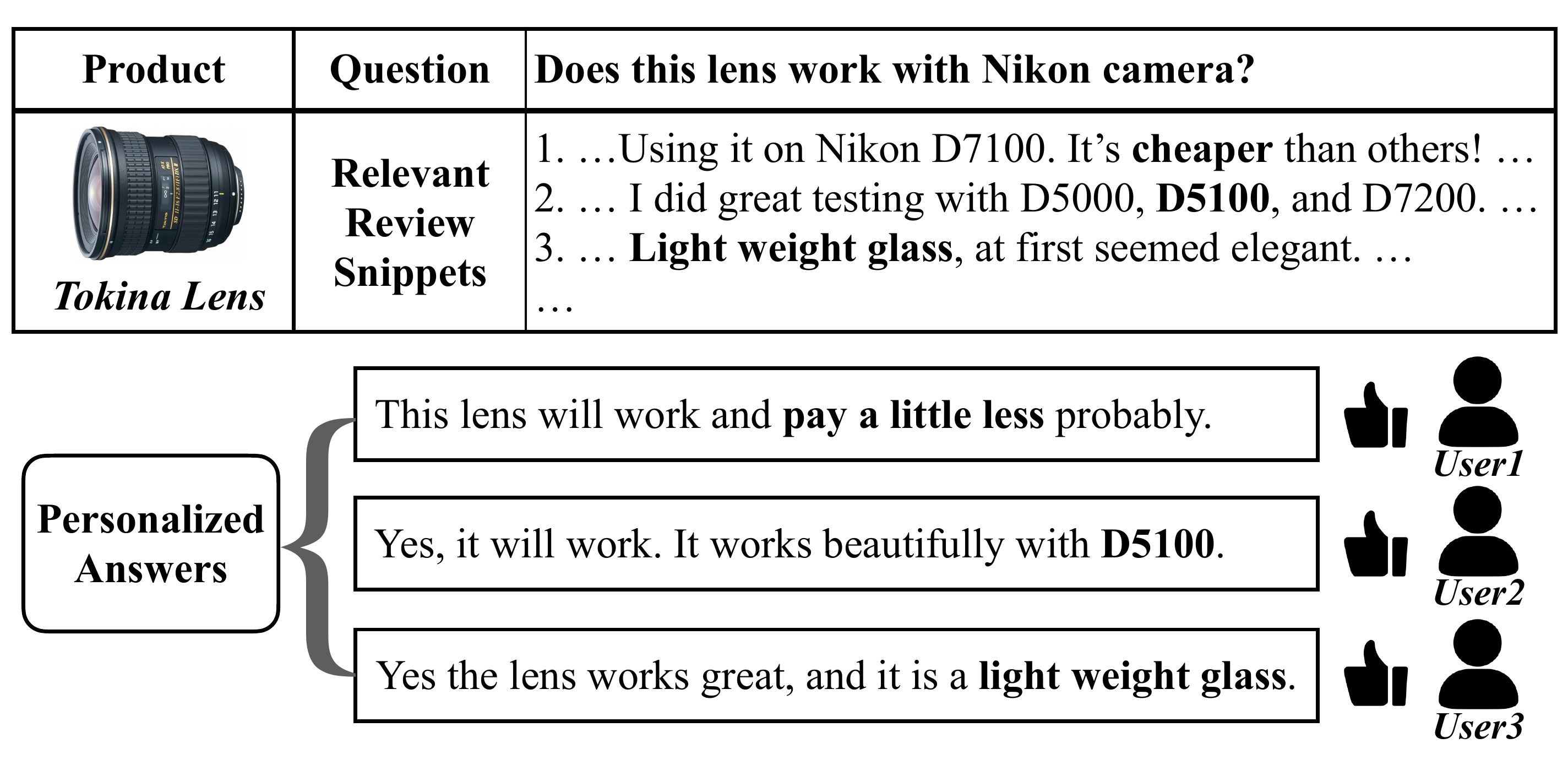}
\caption{A real case from Amazon QA and Review datasets~\cite{icdm16-amazon,emnlp19-amazon-review}. Personalized answer generation is motivated by that different customers may find different answers more helpful, in terms of their own preferences towards product aspects or information needs.} 
\label{example}
\end{figure}

Meanwhile, different customers typically have certain preferences over product aspects or information needs~\cite{kdd19-description,www19-tips}, leading to various expectations for the provided answers. 
As shown in a real case depicted in Figure~\ref{example}, for the same question about the product \textit{Tokina Lens}, different customers may find different answers helpful to themselves. For example, \textit{User1} may focus more on the price of the product, while \textit{User3} cares more about the weight. Besides, \textit{User2} seeks information about a specific camera model. 
This example shows that customers have various preferences and information needs, and the generated answers should take into consideration such diversity. 
Therefore, it is necessary to obtain various information that different customers are interested in from massive product-related data for enriching the provided answers in PQA. This requires the ability of comprehensive user preference modeling, rather than targeting at a single perspective of personalization. 
Due to the nature of E-Commerce platforms, there are a  considerable amount of valuable user-profiling data for user modeling, including user-generated content (UGC), user-item interaction, browsing and clicking history. This leads to extensive studies on personalization in E-Commerce scenario, such as personalized recommendation~\cite{wsdm20-recsys}, review generation and summarization~\cite{acl18-review-gen,emnlp19-review-gen,aaai19-review-summ}, product description generation~\cite{kdd19-description,cikm19-description}, etc. 
To achieve comprehensive user modeling, we leverage historical UGC (including user-posted reviews, answers, and questions) to model multi-perspective user preferences for personalizing the generated answer in PQA.

There are two main challenges that need to be tackled for modeling multi-perspective user preference for PQA: (i) How to obtain user preferences from historical UGC in terms of different perspectives? 
Motivated by the user study in Figure~\ref{user_study}, we investigate approaches to capture users' experience knowledge and preferred product aspects from their historical UGC. Historical UGC can not only be regarded as personal knowledge bases that contain users' experience knowledge, but also serve as the corpus for topic modeling of different product aspects. 
(ii) How to aggregate the multi-perspective user preference information for answer generation in PQA? 
Existing works on personalized text generation typically focus on modeling a specific perspective of user preference, such as learning user features~\cite{kdd19-description,acl19-comment,cikm19-description} or imitating user writing styles~\cite{aaai19-review-summ,www19-tips,emnlp19-review-gen}, by adopting pre-defined user categories or relying on fixed frequency-based user vocabulary for user preference modeling. As for personalized answer generation in PQA, these approaches may fall short of revealing diverse and dynamic preferences from person to person. And it is likely to produce stereotype preference, such as dividing customers into different groups or based on gender or age. 
Instead, it is required the capability of selectively summarizing the most important persona information when modeling multi-perspective user preference.

In this work, we propose a novel \textbf{P}ersonalized \textbf{A}nswer \textbf{GE}neration method (\textbf{PAGE}) that explores historical user-generated content to comprehensively model multi-perspective user preference for PQA. 
We first retrieve question-related user history based on the relevancy to the given question, and leverage such user history content to incorporate valuable personal historical information as external knowledge for \textit{knowledge-level} preference modeling. 
Then we employ Gaussian Softmax distribution Model (GSM) based latent user preference modeling to capture \textit{aspect-level} user preference. This enables us to learn individual preference information as well as provide explicit user-preferred product aspects. 
Meanwhile, a preference-based user vocabulary is dynamically built for providing answers with specific expressions that reflect user preference in \textit{vocabulary-level}. Finally, we develop a persona-aware pointer network to summarize multi-perspective personal information for generating personalized answers, including \textit{knowledge-level}, \textit{aspect-level}, and \textit{vocabulary-level} user preferences. 
Experimental results on real-world E-Commerce QA datasets show that the proposed method outperforms existing methods on both answer generation and personalization in PQA, by comprehensively taking into account multi-perspective user preference.

In summary, our main contributions are as follows:
\begin{itemize}
    \item We investigate a new problem, the personalization of answer generation in E-Commerce, which aims at exploring personal preference from user-generated data to generate personalized answers for different customers. 
    \item We propose a novel \textbf{P}ersonalized \textbf{A}nswer \textbf{GE}neration (\textbf{PAGE}) method with multi-perspective preference modeling to comprehensively model user preference at knowledge-, aspect-, and vocabulary-level from textual personal data. 
    \item We propose a persona-aware pointer-generator network that bridges neural topic modeling and pointer network to dynamically capture user preference and learn a preference-based user vocabulary for personalized text generation.
    \item A real-world E-Commerce QA dataset with historical UGC is constructed for the personalized answer generation studies in PQA\footnote{Available at \url{https://github.com/dengyang17/TOIS-PAGE}}. The proposed method outperforms existing methods on real-world E-Commerce QA datasets and effectively generates informative, diverse, and persona-aware answers. 
\end{itemize}

The rest of the article is organized as follows. Section~\ref{section2} reviews related works.
Section~\ref{section3} introduces the problem definition of the personalized answer generation task in E-Commerce scenario. 
The proposed method for personalized answer generation is described in Section~\ref{section4}. 
Experimental details and results are discussed in Sections~\ref{section5} and~\ref{section6}, respectively. 
Finally, we draw a conclusion about the work and provide some possible directions for future works in Section~\ref{section7}.
\section{Related Works}\label{section2}
This work focuses on a new problem, namely the personalization of answer generation for product-related question answering, which is closely related to the studies of Product Question Answering, Open-domain Generative QA, Personalized Text Generation, and Personalization with UGC in E-Commerce. 

\subsection{Product Question Answering} 
Existing PQA studies typically fall into two problem settings, which are based on the assumption that candidate user-posted answers exist or not. Those PQA studies assuming the existence of candidate answers basically follow the traditional CQA problem setting~\cite{kablstm,tois-dy} to rank a list of user-posted answers~\cite{wsdm18-answer-sel,www19-pqa,sigir20-answer-sel}. \citet{wsdm18-answer-sel} study transfer learning techniques to leverage cross-domain knowledge for E-Commerce answer ranking. \citet{kdd19-answer-sel} and \citet{sigir20-answer-sel} exploit product reviews as external knowledge to enhance the answer ranking in PQA. Besides, \citet{www2020-answer-help} follow previous review helpfulness prediction studies~\cite{www19-review-help1,www19-review-help2} and formulate the answer helpfulness prediction task for finding helpful answers from a set of candidate answers. \citet{answerfact} focus on the factoid questions in PQA and study as a fact checking problem. 
The other group of PQA studies exploits product reviews as the supporting facts to automatically produce answers for those unanswered questions~\cite{www16-amazon-qa,icdm16-amazon}. These works can be further categorized into retrieval-based methods and generation-based methods. Retrieval-based methods~\cite{emnlp12-answer-pred,kdd19-answer-sel,aaai19-answer-sel} aim at selecting review sentences as the answer by measuring the semantic relevance between the question and reviews. 
Text generation methods~\cite{wsdm19-answer-gen-chen,wsdm19-answer-gen-gao,tois21,oaag} employ text generation techniques to generate natural sentences from the relevant product reviews for answering the given question. \citet{wsdm19-answer-gen-chen} propose a gated attention mechanism to capture the question-related information from the reviews for denoising in answer generation. \citet{wsdm19-answer-gen-gao} propose an adversarial learning framework to identify whether the generated answer matches the facts, which is further enhanced by a review-reasoning module and prototype-editing~\cite{tois21}. 
\citet{sigir21-pqa} propose a review-attribute heterogeneous graph neural network to model the logical relation of multi-type text in PQA. 
However, existing studies neglect an important characteristic of PQA, i.e., personalization. 

\subsection{Open-domain Generative QA}
Different from extractive QA~\cite{squad,tois-dy,sigir20-dy}, which aims to select sentences or extract certain text spans from the source document to answer the given question, generative QA~\cite{msmarco,eli5} generates natural sentences as the answer. In factoid QA, some latest works~\cite{emnlp18-answer-gen1,emnlp18-answer-gen2} adopt abstractive summarization methods, based on pointer-generator framework~\cite{acl17-pointer}, to generate an abstractive answer. \citet{acl19-answer-gen} propose a multi-style abstractive summarization model for generative QA with multi-task learning. \citet{emnlp19-answer-gen} incorporate external knowledge to generate a natural answer. 
Recently, pre-trained language models~\cite{nips20-rag,fid} are leveraged to tackle the open-domain generative factoid QA with open-retrieval from large-scale knowledge bases, e.g., Wikipedia.
On the other hand, several efforts have been made on tackling non-factoid generative QA over supporting documents, which mainly focus on ``how'' or ``why'' type questions \cite{wikihowqa,why-question,msg} or aim at generating a conclusion or explanation for the concerned question \cite{eli5,conclusion-ans}.
In E-Commerce scenario, the answer to the given question is often not unique, since it is expected to meet the diverse personal interests and information needs of different users.

\subsection{Personalized Text Generation}\label{sec:related-pergen}
In recent years, personalized textual content generation has attracted research interest from various domains. Personalized dialogue generation~\cite{acl18-dialog} is one of the most active research topics in personalized text generation, which aims at generating persona-aware responses in multi-turn conversation. They mainly focus on exploiting the prerequisite personal textual information to control the persona-consistency~\cite{acl2020-dialog,ecai20-dialog} and enhance the diversity in response generation~\cite{ijcai19-dialog,acl20-dialog-metrics}. Another widely-studied area is personalized review generation~\cite{acl18-review-gen,www20-review-gen,cikm20-review-gen} and summarization~\cite{aaai19-review-summ,www19-tips}. \citet{acl18-review-gen} and \citet{aaai19-review-summ} leverage specific user embedding approaches to incorporate personal information into review generation and summarization, respectively. \citet{emnlp19-review-gen} and \citet{www19-tips} combine autoencoder and discriminator and employ user vocabulary to imitate user writing styles when generating or summarizing product reviews, respectively. Besides, there are some new applications on personalized text generation, including personalized product description generation~\cite{kdd19-description,cikm19-description}, recommendation explanation generation~\cite{recsys19-justification,emnlp19-amazon-review,tist-reason}, personalized recipe generation~\cite{emnlp19-recipe}, etc.
The personalization becomes more appealing in PQA, since there are various perspectives of user preferences that need to be involved in the provided answers.

\subsection{Personalization with UGC in E-Commerce}

As a user-centric application, E-Commerce platforms preserve a considerable amount of valuable user-profiling and user history data for modeling diverse user preferences or behaviors. 
Such user data includes user-generated content (UGC), user-item interaction, browsing and clicking history, etc. 
This leads to extensive studies on different tasks with personalization in E-Commerce scenario, including the personalized text generation studies that we have mentioned in Section~\ref{sec:related-pergen}, like review generation and summarization~\cite{acl18-review-gen,emnlp19-review-gen,aaai19-review-summ}, product description generation~\cite{kdd19-description,cikm19-description}.
One of the most extensively studied research topics involving personalization in E-Commerce is the personalized recommendation~\cite{dl-rec,wsdm20-recsys}, which typically explores the utilities of user-item interaction data to model the user preferences towards different items. 
Recently, several attempts~\cite{sigir20-review4rec,cikm20-review4rec,tois19-review4rec} have been made on leveraging UGC data, such as reviews, for personalized recommendation. The UGC data can not only be served as a measurement of semantic relevance for user preference modeling~\cite{ecir21-review4rec}, but also be adopted as the explicit justifications for explainable recommendation~\cite{tois19-review4rec-explain}. 
Another trending research area in E-Commerce is the personalized product search~\cite{sigir17-product-search,cikm19-product-search,tois19-product-search}, where the UGC data is also beneficial to the semantic product matching~\cite{review4productsearch}. 
In this paper, we investigate the utilities of UGC data for the personalization in PQA.

\begin{table}
\centering
  \caption{Notations}
  \begin{tabular}{cp{7cm}}
    \toprule
	Notation & Definition  \\
    \midrule
    $\bm{X_q}$ & the embedding matrix of the question\\
    $\bm{X_y}$ & the masked embedding matrix of the answer\\
    $\bm{X_{f_k}}$ & the embedding matrix of the $k$-th fact snippet \\
    $\bm{X_{h_k}}$ & the embedding matrix of the $k$-th history snippet\\
    $\bm{d}$ & the bag-of-words (BoW) representation of the $\mathcal{H}^*$\\
    $\bm{T}$&the global preference embeddings\\
    $\bm{V}$&the global vocabulary embeddings\\
    $\bm{M_q}$ & the question memory vector\\
    $\bm{M_f}$ & the fact memory vector\\
    $\bm{M_h}$ & the persona history memory vector \\
    $\bm{M_t}$ & the persona preference memory vector \\
    $\bm{M_p}$ & the persona vocabulary vector\\
    $\bm{s_t}$ & the decoder output at the $t$-th step\\
  \bottomrule
\label{notation}
\end{tabular}
\label{notation}
\end{table}

\section{Problem Definition}\label{section3}
We present the problem of personalized answer generation for product-related questions. The data consists of users, items, questions, answers, supporting facts (including review snippets, product descriptions, and attributes), and historical user-generated content (including reviews, answers, questions). The whole training corpus is denoted by $\mathcal{D} = \{\mathcal{U}, \mathcal{I}, \mathcal{Q}, \mathcal{A}, \mathcal{F}, \mathcal{H}\}$, where $\mathcal{U}$ and $\mathcal{I}$ are the sets of users and items respectively, $\mathcal{Q}$ and $\mathcal{A}$ are the sets of questions and corresponding user-written answers respectively, $\mathcal{F}_i$ is the set of supporting facts for the item $i$ in $\mathcal{I}$, and $\mathcal{H}_j$ is the set of historical UGC from the user $j$ in $\mathcal{U}$. Due to the large number of reviews, we extract top-$k$ snippets $\mathcal{F}^*_i$ and $\mathcal{H}^*_j$ from $\mathcal{F}_i$ and $\mathcal{H}_j$, which are most relevant to the given question based on information retrieval (IR) techniques.

With the above definitions, our problem can be formally defined as: Given a product-related question $q\in\mathcal{Q}$ concerning the item $i$, we aim to automatically generate a personalized and informative answer $y$ for the specific user $j$, based on the relevant supporting facts $\mathcal{F}^*_i=\left\{f_1,f_2,...,f_k\right\}_i$ and historical UGC $\mathcal{H}^*_j=\left\{h_1,h_2,...,h_k\right\}_j$. For simplicity, we omit $i$ and $j$ in the following notations. 
Apart from the above notations, the notations used in the methodology are listed in Table~\ref{notation}.

\section{Personalized Answer Generation}\label{section4}

\begin{figure*}
\centering
\includegraphics[width=\textwidth]{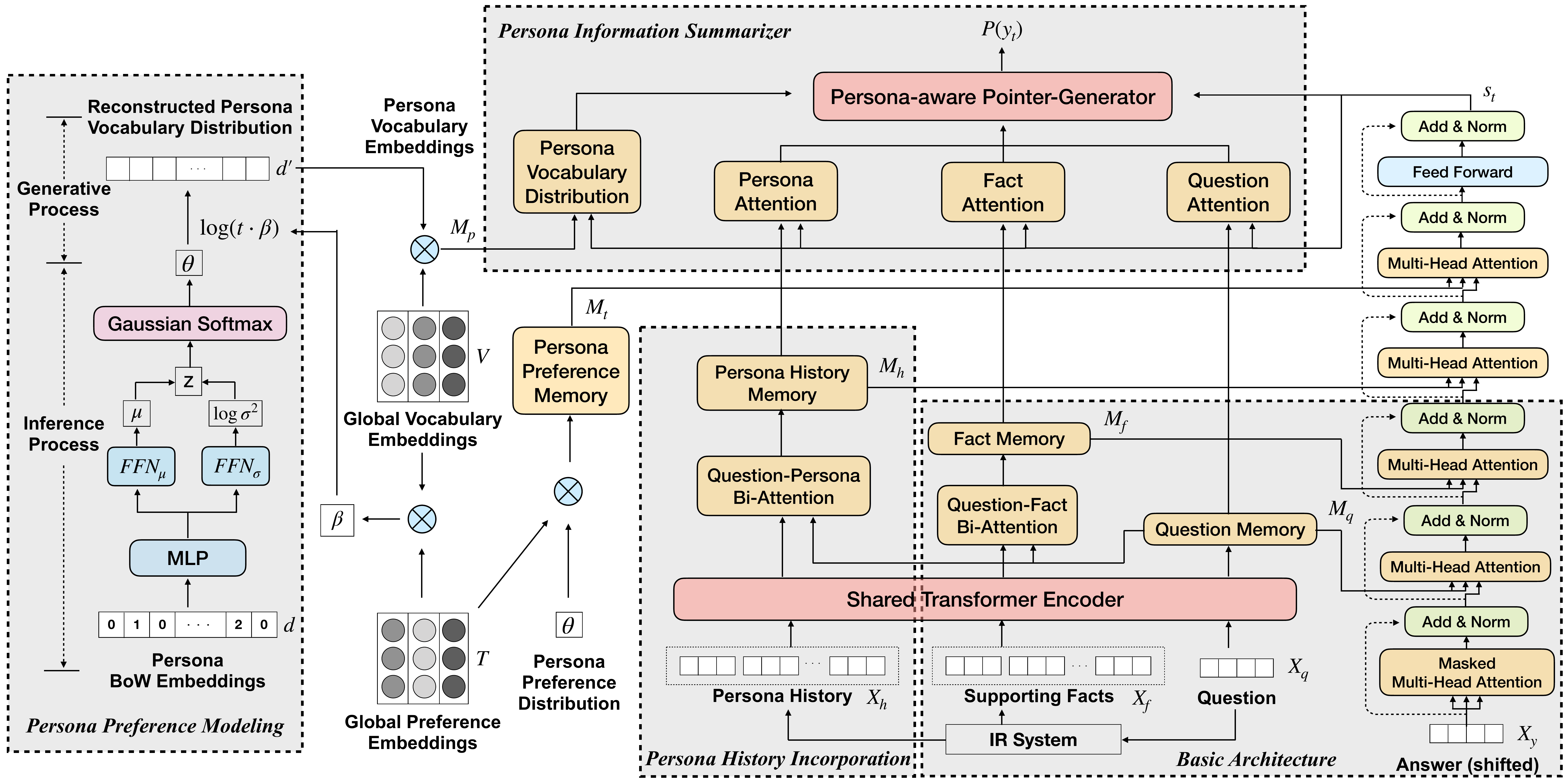}
\caption{Overview of the proposed method PAGE, including four components: (1) Basic Encoder-decoder Architecture, (2) Persona History Incorporation, (3) Persona Preference Modeling, and (4) Persona Information Summarizer.}
\label{method}
\end{figure*}

As shown in Figure~\ref{method}, the proposed method, PAGE, consists of four components: 
\begin{itemize}
    \item (1) \textit{Basic Encoder-decoder Architecture} is the base model for general answer generation without personalization, which takes the question $q$ and the retrieved supporting facts $\mathcal{F}^*$ as input and outputs the decoded vocabulary probability distribution $P_v(a)$;
    \item (2) \textit{Persona History Incorporation} module incorporates historical UGC as external knowledge for knowledge-level user preference modeling, which takes the retrieved historical UGC $\mathcal{H}^*$ as input and outputs the persona history memory vector $M_h$;
    \item (3) \textit{Persona Preference Modeling} module employs neural topic model to capture latent aspect-level user preference as well as dynamically provide preference-based user vocabulary, which takes the BoW representation $d$ of the $\mathcal{H}^*$ as input, and outputs the persona preference memory vector $M_t$ and the persona vocabulary vector $M_p$;
    \item (4) \textit{Persona Information Summarizer} module summarizes the multi-perspective user preference information for generating personalized answers, which aggregate the multi-perspective information to calculate the final vocabulary probability distribution $P(y_t)$.
\end{itemize}
   
\subsection{Basic Encoder-decoder Architecture}\label{sec:encdec}
We first describe the base model that only performs general answer generation without the consideration of personal information.
We combine the Transformer~\cite{nips17-transformer} and the BiDAF (Bidirectional Attention Flow)~\cite{iclr17-bidaf} as the basic encoder-decoder architecture. 

\subsubsection{\textbf{Self-Attentive Encoder Layer}}
For the encoder, we use a shared self-attentive encoder to compute the representations of the question, supporting facts and history content separately. Each self-attentive encoder layer consists of three components: (i) The layer normalization, which is defined as LayerNorm$(\cdot)$. (ii) The multi-head attention, which is defined as MultiHead$(\bm{Q}, \bm{K}, \bm{V})$, where $\bm{Q},\bm{K},\bm{V}$ are query, key, and value, respectively. (iii) The feed-forward network with ReLU activation, which is defined as FFN$(\cdot)$. And the encoder is composed of a stack of $N_e$ identical layers. Take the first encoder layer for question encoding for example:
\begin{gather}
    \bm{C_*} = \text{MultiHead}(\bm{X_*},\bm{X_*},\bm{X_*}), \\
    \bm{O_*}=\text{LayerNorm}(\text{FFN}(\bm{C_*}) + \bm{X_*}),
\end{gather}
where $*\in\{q,f_1,f_2,...,f_k\}$ represents the word sequence of the question and each supporting fact snippets, $\bm{X_*}$ denotes their embeddings, and $\bm{O_*}\in\mathbb{R}^{L_*\times d_h}$ is the output of the encoder layer. $L_*$ and $d_h$ denote the length of the corresponding sequence and the dimension of the hidden layer, respectively. 

\subsubsection{\textbf{Bi-Attention Layer}}
After obtaining the encoded representation for all the input sequences, we apply a bidirectional attention layer (BiAttention)~\cite{iclr17-bidaf} to refine the question-related information in the supporting facts. For each supporting fact snippet $f_k$, we derive a similarity matrix $\bm{U_{f_k}}\in\mathbb{R}^{L_q\times L_f}$ between the question and snippet:
\begin{equation}
    \bm{U_{f_k}^{(ij)}} = \bm{\omega_U}^\intercal[\bm{O_q^{(i)}};\bm{O_{f_k}^{(j)}};\bm{O_q^{(i)}}\circ \bm{O_{f_k}^{(j)}}],
\end{equation}
where $\bm{\omega_U}\in\mathbb{R}^{3d_h\times 1}$ is a trainable weight vector, $\circ$ denotes element-wise multiplication, and [;] denotes the concatenate operation. Then, we can obtain the row and column normalized similarity vectors as the question-to-fact and fact-to-question attention weights, $\bm{A_{f_k}}$ and $\bm{B_{f_k}}$, respectively:
\begin{gather}
    \bm{A_{f_k}} = \text{softmax}(\bm{U_{f_k}}), \\
    \bm{B_{f_k}} = \text{softmax}(\bm{U_{f_k}}^\intercal).
\end{gather}
After being weighted by the attentions, the attended question and fact representations will be:
\begin{gather}
    \widetilde{\bm{O_q}} = \bm{A_{f_k}}^\intercal \bm{O_q}, \\ \widetilde{\bm{O_{f_k}}} = \bm{B_{f_k}}^\intercal \bm{O_{f_k}}.
\end{gather}
And we use a simple concatenation to get the final fact memory representation $\bm{M_{f_k}}\in\mathbb{R}^{L_f\times 4d_h}$:
\begin{equation}
\begin{split}
    \bm{M_{f_k}} = & ~\text{BiAttention}(\bm{O_{f_k}}, \bm{O_q}) \\
    = & ~ [\bm{O_{f_k}};\widetilde{\bm{O_q}};\bm{O_{f_k}}\circ\widetilde{\bm{O_q}};\bm{O_{f_k}}\circ\widetilde{\bm{O_{f_k}}}].
\end{split}
\end{equation}

\subsubsection{\textbf{Decoder Layer}}
Similar to the encoder, we adopt a stack of $N_d$ Transformer decoder layers that contain multi-head attention and feed-forward network as the decoder. The difference is that the decoder employs not only the source text information, i.e., the question, but also the information from supporting facts to generate the answer. For each decoder layer:
\begin{gather}
    \bm{H_y}=\text{MultiHead}(\bm{X_y},\bm{X_y},\bm{X_y}),\\
    \bm{H_q}=\text{MultiHead}(\bm{H_y},\bm{O_q},\bm{O_q}),\\
    \bm{H_f}=\text{MultiHead}(\bm{H_q},\bm{M_f},\bm{M_f}),\\
    \bm{O_{dec}} = \text{FFN}(\bm{H_f}),
\end{gather}
where $\bm{X_y}$ is the masked answer embedding, $\bm{M_f}$ is the concatenated outputs of all the supporting fact snippets $\bm{M_f} = [\bm{M_{f_1}},\bm{M_{f_2}},...,\bm{M_{f_k}}]\in\mathbb{R}^{kL_f\times d_h}$, and $\bm{O_{dec}}$ is the decoder output. For simplicity, we omit the layer normalization and residual add operation~\cite{nips17-transformer} in the equation. 

Finally, the decoder output $\bm{O_{dec}}$ is projected into the vocabulary with a softmax function to obtain the vocabulary distribution:
\begin{equation}
    P_v(a)=\text{softmax}(\bm{W_v}\bm{O_{dec}}+\bm{b_v}),
\end{equation}
where $\bm{W_v}\in\mathbb{R}^{d_h\times|V|}$ and $\bm{b_v}\in\mathbb{R}^{|V|}$ are the projection parameters to be learned, and $|V|$ is the vocabulary size.

\subsection{Persona History Incorporation}\label{sec:phi}

\begin{figure*}
\centering
\includegraphics[width=0.8\textwidth]{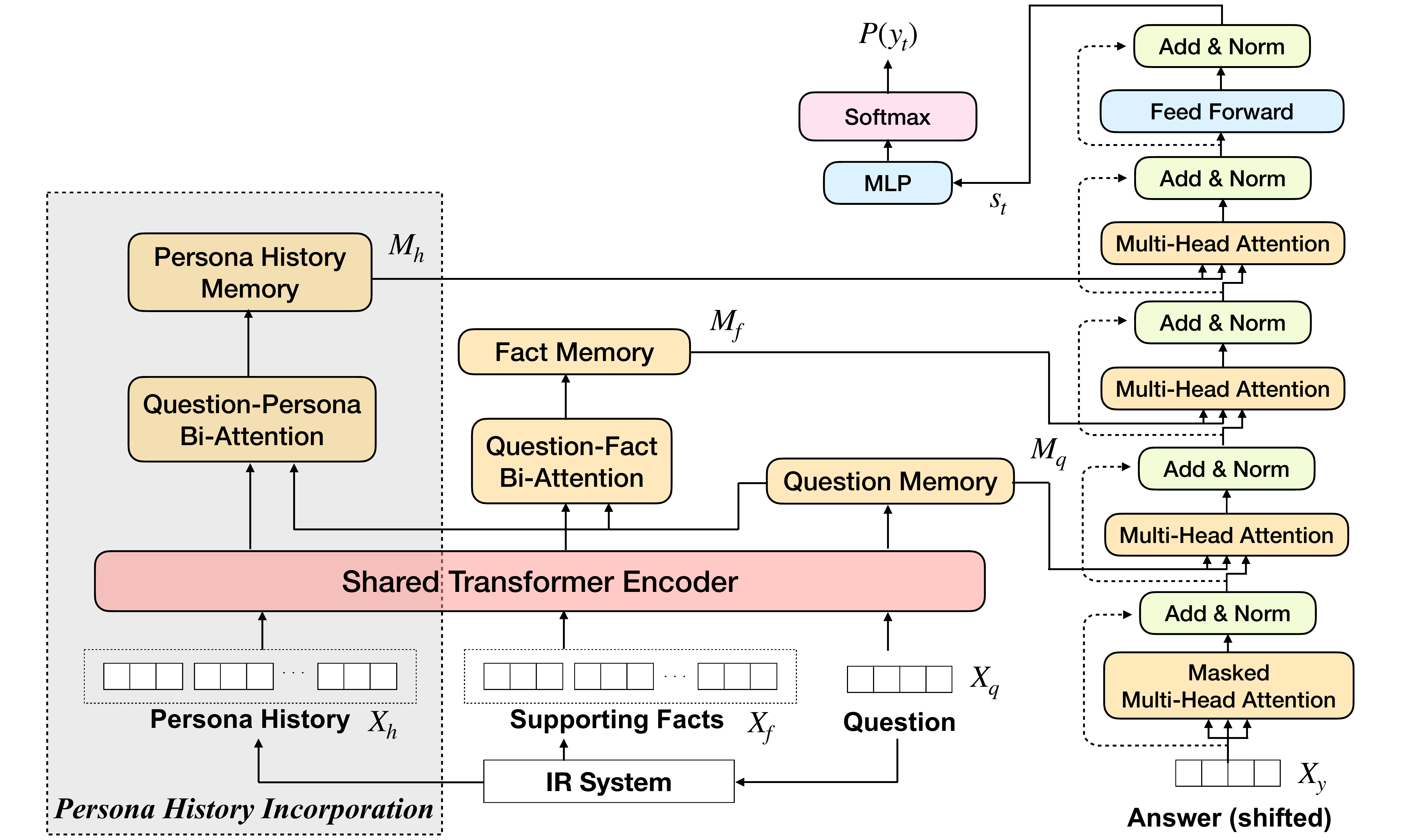}
\caption{Basic Encoder-decoder Architecture with Persona History Incorporation (PHI). An additional Multi-Head Attention layer is added into the decoder for incorporating the persona history information. }
\label{phi}
\end{figure*}

In order to generate a personalized and informative answer for a specific user, it is insufficient to only exploit the general supporting facts. In real-world applications, customers usually expect to receive some practical suggestions or seek certain information needs according to their own experience. For example, users would like to know about the specific answer corresponding to certain camera model as the example in Figure~\ref{example}, or under particular situations. And this prior knowledge can be discovered from user history. 

To this end, we leverage question-related historical user-posted content $\mathcal{H}^*$ as external knowledge for incorporating their personal history. 
We employ the same shared self-attentive encoder and bi-attention mechanism to encode the retrieved historical content snippets:
\begin{gather}
    \bm{C_{h_k}} = \text{MultiHead}(\bm{X_{h_k}},\bm{X_{h_k}},\bm{X_{h_k}}),\\
    \bm{O_{h_k}}=\text{LayerNorm}(\text{FFN}(\bm{C_{h_k}}) + \bm{X_{h_k}}),\\
    \bm{M_{h_k}} = \text{BiAttention}(\bm{O_{h_k}}, \bm{O_q}),
\end{gather}
where $\bm{M_{h_k}}$ is the encoded persona history memory vector. Then, we can incorporate the user history information into the decoding phase for knowledge-level user preference modeling. 
The basic encoder-decoder architecture with Persona History Incorporation (PHI) module is depicted in Figure~\ref{phi}. 

\subsection{Persona Preference Modeling}\label{sec:ppm}

\begin{figure*}
\centering
\includegraphics[width=0.9\textwidth]{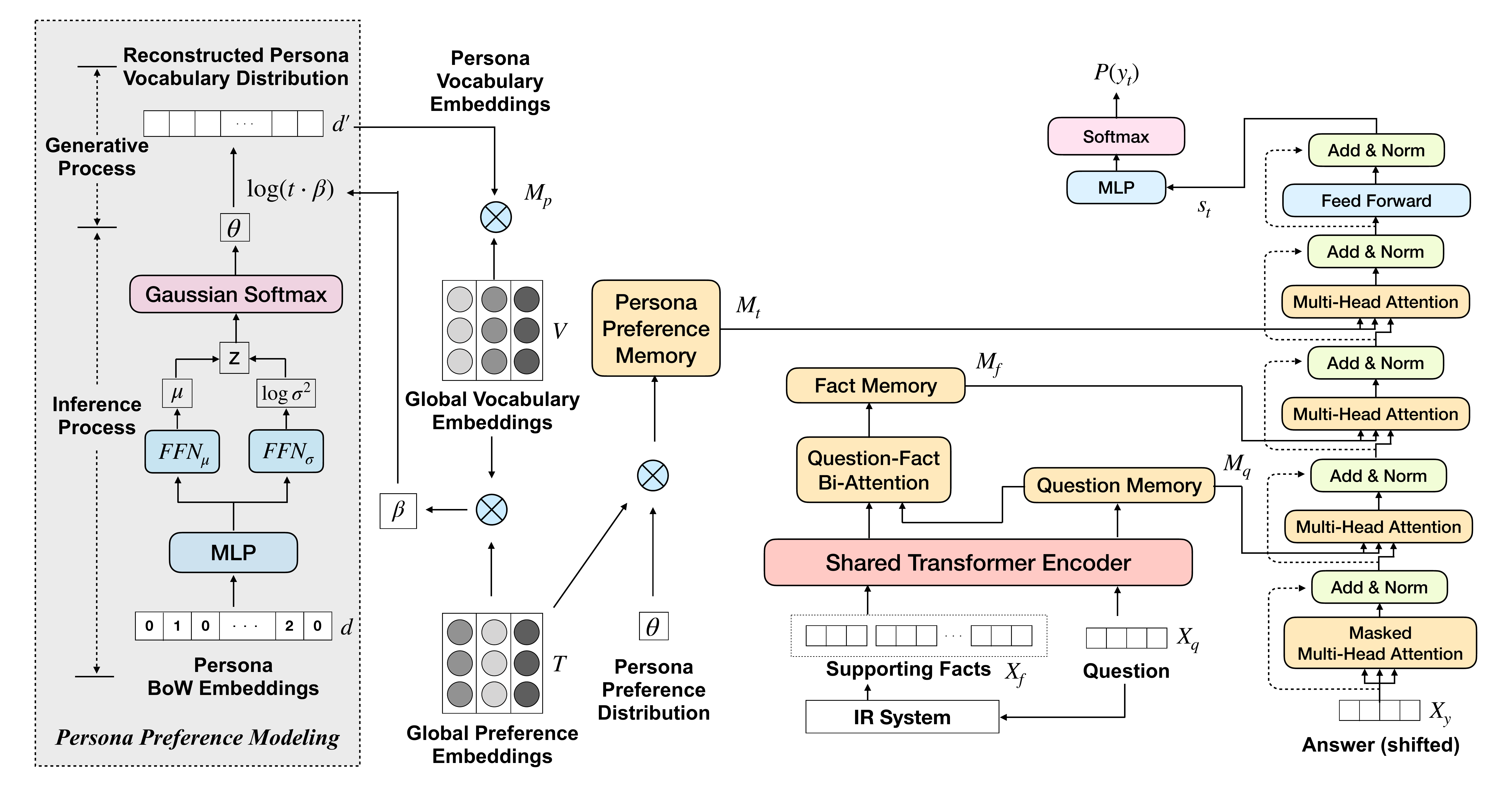}
\caption{Basic Encoder-decoder Architecture with Persona Preference Modeling (PPM). An additional Multi-Head Attention layer is added into the decoder for incorporating the persona preference information. }
\label{ppm}
\end{figure*}

As discussed in Section~\ref{sec:intro}, only modeling knowledge-level user preference is insufficient for personalized answer generation. Different users also may have their own latent aspect-level preference or explicit vocabulary-level preference towards various answers. Instead of using pre-defined user categories or fixed frequency-based user vocabulary, persona preference model (PPM) module employs neural topic model to capture latent aspect-level user preference as well as dynamically provide preference-based user vocabulary. 
The basic encoder-decoder architecture with Persona Preference Modeling (PPM) module is depicted in Figure~\ref{ppm}. 

To be specific, Gaussian Softmax distribution model (GSM)~\cite{icml17-gsm} is adopted for unsupervised topic modeling, which is based on variational autoencoder (VAE)~\cite{iclr14-vae}. 
Formally, similar to persona history incorporation module, PPM also adopts a set of question-related historical UGC $\mathcal{H}^*$ as the input. The difference is that PPM processes the collection of each user's UGC into bag-of-words (BoW) representation $\bm{d}$, instead of using the original natural sentences. The purpose is to enable the interpretation of latent aspect-level preference modeling as well as provide the foundation of vocabulary-level preference modeling. 
In the neural variational inference process, the input $\bm{d}$ is first encoded into continuous Gaussian variables $\bm{\mu}$ and $\bm{\sigma}$ via a multi-layer perceptron:
\begin{gather}
\mu = \text{FFN}_{\bm{\mu}}(\text{FFN}_d(\bm{d})), \\
\text{log}(\bm{\sigma}^2) = \text{FFN}_{\sigma}(\text{FFN}_d(\bm{d})).
\end{gather}

The prior $p(\bm{x}) = \mathcal{N}(\bm{\mu},{\bm{\sigma}}^2)$ is a diagonal Gaussian distribution with mean $\bm{\mu}$ and $\bm{\sigma}^2$ as the diagonal of the covariance matrix. By using the Gaussian prior distribution, we can build an unbiased gradient estimator for the variational distribution with the re-parameterization trick~\cite{iclr14-vae}. Then, we use latent variable $\bm{\theta}\in\mathbb{R}^K$, which is derived by the Gaussian Softmax operation over the prior distribution $\bm{x}$, to denote the probability distribution over the aspect-level user preference cluster:
\begin{gather}
\bm{x}\sim \mathcal{N}(\bm{\mu},\bm{{\sigma}}^2),\\
\mathbf{\bm{\theta}} = \text{softmax}(\bm{W_x}\bm{x}+\bm{b_x}),
\end{gather}
where $K$ is the predefined number of aspect clusters, $\bm{W_x}$ and $\bm{b_x}$ are the weights and bias of linear transformation to be learned. In practise, we reparameterise $\bm{x}=\bm{\mu}+\epsilon\cdot\bm{\sigma}$ with the sample $\epsilon\in\mathcal{N}(0,I)$. We use $q(\bm{\theta}|\bm{d})$ to denote the variational inference process approximating the true posterior $p(\bm{\theta}|\bm{d})$.

We denote the aspect assignment as $z_n$ for the observed word $w_n$. A neural network is used to parameterise the multinomial aspect distribution, and the generative process for $\bm{d}$ is:
\begin{gather}
z_n \sim \text{Multi}(\bm{\theta}), ~ \text{for} ~ n \in [1,|V_{bow}|],\\
w_n \sim \text{Multi}(\bm{\beta_{z_n}}), ~ \text{for} ~  n \in [1,|V_{bow}|],
\end{gather}
where $\bm{\beta_{z_n}}$ denotes the aspect distribution over words given aspect assignment $z_n$, and $V_{bow}$ is the vocabulary for BoW model. Then, the marginal likelihood is computed by:
\begin{equation}\label{eq:marginal_likelihood}
    p(\bm{d}|\bm{\mu},\bm{\sigma},\bm{\beta}) = \int_{\bm{\theta}}p(\bm{\theta}|\bm{\mu},\bm{\sigma})\prod_n\sum_{z_n}p(w_n|\bm{\beta_{z_n}})p(z_n|\bm{\theta})d\bm{\theta}.
\end{equation}

In order to explicitly model the user preference, the aspect distribution over words $\bm{\beta}$ is constructed by the product of the global preference embeddings $\bm{T}\in\mathbb{R}^{K\times d_h}$ and the global vocabulary embeddings $\bm{V}\in\mathbb{R}^{|V_{bow}|\times d_h}$, instead of being regarded as a single parameter matrix:
\begin{equation}
    \bm{\beta} = \text{softmax}(\bm{T}\bm{V}^\intercal),
\label{eq:beta}
\end{equation}
where $\bm{T}$ and $\bm{V}$ are randomly initialized embeddings to be learned. Therefore, $\bm{\beta}\in\mathbb{R}^{K\times|V_{bow}|}$ can be regarded as the semantic similarity matrix between aspects and words. Then we can decouple the aspect-level and vocabulary-level user preference easily. We reconstruct a new vector $\bm{d'}$ by
\begin{equation}
    \bm{d'}=\bm{\theta}\cdot\bm{\beta},
\end{equation}
which is served as the user vocabulary distribution based on the persona preference modeling. 

The aspect distribution $\bm{\theta}$ and the vocabulary distribution $\bm{d'}$ are combined with the global preference embeddings $\bm{T}$ and the global vocabulary embeddings $\bm{V}$ to build the persona preference memory vectors $\bm{M_t}$ and persona vocabulary vectors $\bm{M_p}$ for each user:
\begin{gather}
    \bm{M_t}=\bm{\theta}\cdot \bm{T}, \\
    \bm{M_p}=\bm{d'}\cdot \bm{V}.
\end{gather}
The persona preference memory vectors $\bm{M_t}$ are further integrated into the answer decoding phase for integrating aspect-level user preference, while the persona vocabulary vectors $\bm{M_p}$ are used to enhance the generated answers with user-preferred language styles.

For PPM, the objective function is defined based on the negative variational lower bound~\cite{icml17-gsm} according to Equation (\ref{eq:marginal_likelihood}):
\begin{equation}
\begin{split}
    \mathcal{L}_{PPM}=&-\mathbb{E}_{q(\bm{\theta}|\bm{d})}\left[\sum_{n=1}^{|V_{bow}|}\text{log}\sum_{z_n}\left[p(w_n|\bm{\beta_{z_n}})p(z_n|\bm{\theta})\right]\right]\\
    &+\mathcal{D}_{KL}[q(\bm{\theta}|\bm{d})||p(\bm{\theta}|\bm{\mu},\bm{\sigma})],
\end{split}
\end{equation}
where, given a sampled $\bm{\hat{\theta}}$, the latent variable $z_n$ can be integrated out as:
\begin{equation}
    \text{log}p(w_n|\bm{\beta},\bm{\hat{\theta}}) = \text{log}\sum_{z_n}\left[p(w_n|\bm{\beta_{z_n}})p(z_n|\bm{\hat{\theta}})\right]=\text{log}(\bm{\hat{\theta}}\cdot\bm{\beta}).
\end{equation}
Then, the objective function of PPM can be finalized as:
\begin{equation}\label{eq:ppm}
    \mathcal{L}_{PPM}=-\sum_{n=1}^{|V_{bow}|}\text{log}p(w_n|\bm{\beta},\bm{\hat{\theta}})+\mathcal{D}_{KL}[q(\bm{\theta}|\bm{d})||p(\bm{\theta}|\bm{\mu},\bm{\sigma})].
\end{equation}

\subsection{Persona Information Summarizer}\label{sec:pis}
In order to summarize the multi-perspective user preference information, we propose a persona-aware pointer generator network to generate answers involving knowledge-level and aspect-level user preference as well as copy words from a preference-aware user vocabulary. 
In specific, the knowledge-level user preference is obtained from the Persona History Incorporation (PHI) module in Section~\ref{sec:phi}, while the aspect-level user preference and the preference-aware user vocabulary are obtained from the Persona Preference Modeling (PPM) module in Section~\ref{sec:ppm}.

\begin{figure*}
\centering
\includegraphics[width=0.7\textwidth]{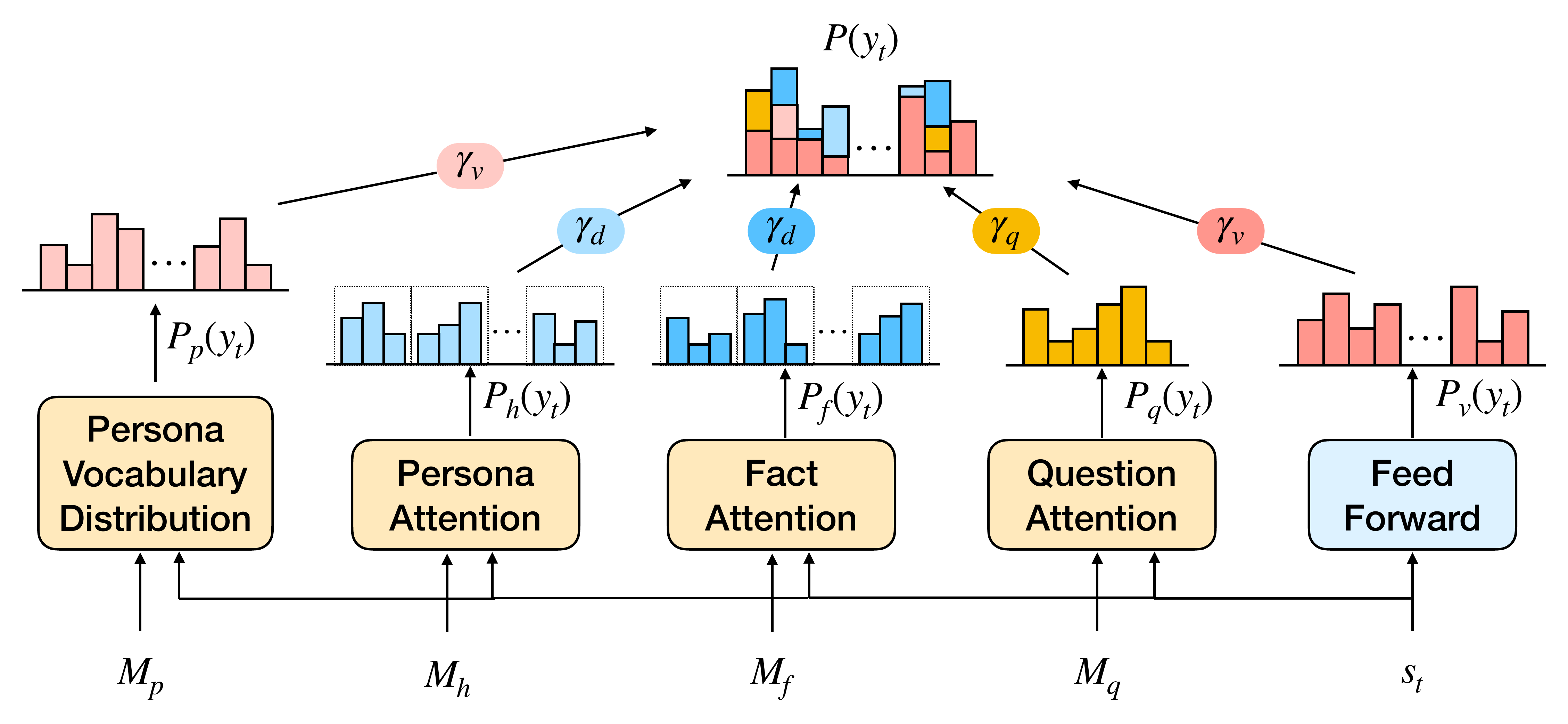}
\caption{Persona Information Summarizer}
\label{psi}
\end{figure*}

After incorporating the knowledge-level and aspect-level user preference information, the decoder layer becomes:
\begin{gather}
    \bm{H_h}=\text{MultiHead}(\bm{H_f},\bm{M_h},\bm{M_h}),\\
    \bm{H_p}=\text{MultiHead}(\bm{H_h},\bm{M_t},\bm{M_t}),\\
    \bm{O_{dec}} = \text{FFN}(\bm{H_p}),
\end{gather}
where $\bm{M_h}$ is the concatenated persona history memory vector of all the user historical content snippets $\bm{M_h} = [\bm{M_{h_1}},\bm{M_{h_2}},...,\bm{M_{h_k}}]\in\mathbb{R}^{kL_h\times d_h}$.

Let $y_t$ and $\bm{s_t}$ be the decoder input and output at the $t$-th step, and $\bm{M_q}=\bm{O_q}$ to be the question memory vector for avoiding the notational clutter. The attention weight for each word in the question, the concatenated supporting facts, historical content, and the learned user vocabulary, i.e., $\bm{\alpha^q_t}$, $\bm{\alpha^f_t}$, $\bm{\alpha^h_t}$, and $\bm{\alpha^p_t}$, are generated by:
\begin{gather}
    \bm{\alpha^q_t} = \text{softmax}(\bm{\omega_q}^\intercal \text{tanh} (\bm{W_{q}} \bm{M_q}+\bm{W_{qs}} \bm{s_t}+\bm{b_{q}})),\\
    \bm{\alpha^f_t} = \text{softmax}(\bm{\omega_f}^\intercal \text{tanh} (\bm{W_{f}} \bm{M_f}+\bm{W_{fs}} \bm{s_t}+\bm{b_{f}})),\\
    \bm{\alpha^h_t} = \text{softmax}(\bm{\omega_h}^\intercal \text{tanh} (\bm{W_{h}} \bm{M_h}+\bm{W_{hs}} \bm{s_t}+\bm{b_{h}})),\\
    \bm{\alpha^p_t} = \text{softmax}(\bm{\omega_p}^\intercal \text{tanh} (\bm{W_{p}} \bm{M_p}+\bm{W_{ps}} \bm{s_t}+\bm{b_{p}})),
\end{gather}
where $\bm{W_*}$, $\bm{W_{*s}}$, $\bm{\omega_*}$ and $\bm{b_{*}}$ ($*\in\{q,f,h,p\}$) are parameters to be learned. The attention weights $\bm{\alpha^*_t}$ are used to compute context vectors $\bm{c^*_t}$ as the probability distribution over the source words:
\begin{gather}
    \bm{c^q_t} = \bm{M_q}^\intercal\bm{\alpha^{q}_t},\quad
    \bm{c^f_t} = \bm{M_f}^\intercal\bm{\alpha^{f}_t},\\
    \bm{c^h_t} = \bm{M_h}^\intercal\bm{\alpha^{h}_t},\quad
    \bm{c^p_t} = \bm{M_p}^\intercal\bm{\alpha^{p}_t}.
\end{gather}

Then, we obtain the attention-based probability distribution by indexing the attention weight of each word in both the question and review to the extended vocabulary:
\begin{gather}
    P_q(y_t) = \sum\nolimits_{i:w_i=w} \bm{\alpha^{q_i}_t},\quad
    P_f(y_t) = \sum\nolimits_{i:w_i=w} \bm{\alpha^{f_i}_t},\\
    P_h(y_t) = \sum\nolimits_{i:w_i=w} \bm{\alpha^{h_i}_t},\quad
    P_p(y_t) = \sum\nolimits_{i:w_i=w} \bm{\alpha^{p_i}_t},
\end{gather}
where $P^q(y_t)$, $P^f(y_t)$, $P^h(y_t)$, and $P^p(y_t)$ denote the attention-based probability distribution for words from the question, supporting fact, historical content, and user vocabulary, respectively.

The final probability distribution of $y_t$ is obtained from five views of word distributions, which incorporate the basic information from the question and supporting facts, the external knowledge from user history, and the preference-based user vocabulary:
\begin{gather}
    P_{all}(y_t) = [P_v(y_t),P_q(y_t),P_f(y_t),P_h(y_t),P_p(y_t)],\\
    \bm{\gamma} = \text{softmax}(\bm{W_{\gamma}}[\bm{s_t}:\bm{c^q_t}:\bm{c^f_t}:\bm{c^h_t}:\bm{c^p_t}] + \bm{b_{\gamma}}),\\
    P(y_t) =  P_{all}(y_t)^\intercal \bm{\gamma},
\end{gather}
where $\bm{W_\gamma}$ and $\bm{b_\gamma}$ are parameters to be learned, $\bm{\gamma}$ is the multi-view pointer scalar to determine the weight of each view of probability distribution. 
The overview of the Persona Information Summarizer is depicted in Figure~\ref{psi}.

\subsection{Training}
The proposed method jointly learns latent preference modeling and answer generation in an end-to-end fashion. Apart from the objective function of PPM formulated in Equation~(\ref{eq:ppm}), the objective function of answer generation is trained to minimize the negative log likelihood:
\begin{equation}
    \mathcal{L}_{QA} = - \frac{1}{T}\sum\nolimits^T_{t=0}\text{log}P(w^*_t),
\end{equation}
where $w^*_t$ is the target word at the $t$-th timestep.

To ensure the diversity of the resulting preference embeddings in PPM, we add a regularization term to the objective function $\mathcal{L}_{PPM}$ to differentiate each preference embedding in the global preference embeddings $\bm{T}$:
\begin{equation}
    \mathcal{R} = ||\bm{T}\cdot \bm{T}^\top - \bm{I}||,
\end{equation}
where $\bm{I}$ is the identity matrix.
Finally, the overall training objective of the proposed method is a weighted sum of $\mathcal{L}_{QA}$, $\mathcal{L}_{PPM}$, and the regularization term $\mathcal{R}$:
\begin{equation}
    \mathcal{L} = \mathcal{L}_{QA} + \lambda_1\cdot(\mathcal{L}_{PPM} + \lambda_2\cdot\mathcal{R}),
\end{equation}
where $\lambda$ are hyper-parameters to balance losses.

\section{Experimental Settings}\label{section5}

\subsection{Dataset Collection}
We evaluate our model on a public real-world E-Commerce dataset, Amazon Question/Answer Dataset~\cite{icdm16-amazon}, 
which contains around 1.4 million product-related questions with multiple answers provided by different users. 
This QA dataset can be combined with Amazon Product Dataset~\cite{emnlp19-amazon-review}, which preserves product reviews and metadata from Amazon. 
Besides, there is another public real-world PQA dataset~\cite{wsdm19-answer-gen-gao} collected from one of the largest Chinese E-Commerce platforms, namely JD. However, this dataset does not contain the user information. Hence we are not able to adopt it for the personalized PQA evaluation.  
Following previous studies on answer generation in E-Commerce~\cite{wsdm19-answer-gen-chen,wsdm19-answer-gen-gao,ijcai19-amazonqa}, we adopt user-written answers as the ground-truth for the experiment. 
However, different from these works, which assume that there is only one ground-truth answer for each question, we need to collect a set of personalized answers for different users towards the same question. 
In fact, the user-written answers can reflect the user preference towards the given question more precisely and truly. 
To this end, following previous review/answer helpfulness prediction studies~\cite{www19-review-help1,www19-review-help2,www2020-answer-help}, we regard those answers, which receive more than two votes as well as more upvotes than downvotes, as a good or helpful answer to be the ground-truth answer for the answerer and the given question. 
Then, we use the User ID of the answerer to fetch his/her historical posted content, including questions, answers, and reviews, as the personal information.

Since there are a large number of reviews and metadata, including the description and features, for each product, we need to extract those supporting facts that contain relevant information for each question. 
Similar to~\citet{wsdm19-answer-gen-chen} and~\citet{ijcai19-amazonqa}, each supporting fact text is chunked into snippets of length 50, or to the end of a sentence boundary. Then for a given question, we adopt BM25 to rank all the supporting fact snippets of the corresponding product and collect the top 5 relevant supporting fact snippets for each question as the model input. The same process is conducted for retrieving the user history that is relevant to the given question.

Each sample in the final dataset contains a question, a reference personal answer, a set of user history snippets, and a set of supporting fact snippets. Three categories with the largest number of samples are adopted, namely \textit{Electronics}, \textit{Home\&Kitchen} and \textit{Sports\&Outdoors}. For each category, we leave 1,000 instances for validation, 2,000 for testing, and the remaining are used for training. The statistics of the dataset are presented in Table~\ref{dataset}.

\subsection{Baseline Methods}
We first compare our model to state-of-the-art generative PQA models that exclude personalization:
\begin{itemize}
    \item \textbf{S2SAR}~\cite{wsdm19-answer-gen-gao}. A method incorporates the review information into Sequence-to-sequence model with attention (S2SA), by concatenating the question and reviews as the source text.
    \item \textbf{RAGE}~\cite{wsdm19-answer-gen-chen}. A state-of-the-art review-driven answer generation method for product-related questions\footnote{\url{https://github.com/WHUIR/RAGE}}. 
    \item \textbf{TFMR}. The basic encoder-decoder architecture described in Section~\ref{sec:encdec} combines the Transformer~\cite{nips17-transformer} and the BiDAF~\cite{iclr17-bidaf} for answer generation.
\end{itemize}

\begin{table}
\centering
\caption{The statistics of datasets}
  \begin{tabular}{c|ccc}
  \toprule
  Dataset & Electronics & Home\&Kitchen & Sports\&Outdoors\\
  \midrule
  \midrule
   \#(Q,A) & 39,478/1K/2K&29,867/1K/2K&12,458/1K/2K\\
    Len(Q)& 28.3/25.1/18.4 &20.0/21.1/18.1 &18.9/20.3/18.8\\
   Len(A)& 53.5/50.7/48.5 &46.1/48.9/43.9 &46.8/48.8/45.5\\
   \#User & 35,546/\textasciitilde1K/\textasciitilde2K & 27,691/\textasciitilde1K/\textasciitilde2K &11,377/\textasciitilde1K/\textasciitilde2K\\
  \#History & 33.8/29.1/33.3 &34.0/38.8/36.2 &27.7/29.3/25.3\\
  \bottomrule
  \end{tabular}
\label{dataset}
\end{table}

In order to validate the effectiveness of our proposed method on generating personalized answers, we adapt several latest personalized text generation methods for comparison as follows:
\begin{itemize}
    \item \textbf{Per-S2SA}~\cite{acl18-dialog} is a persona-based generation baseline that appends historical UGC to the input sequence. Here we further concatenate the supporting facts to the input, namely \textbf{Per-S2SAR}.
    \item \textbf{Per-CVAE}~\cite{ijcai19-dialog}\footnote{\url{https://github.com/vsharecodes/percvae}} and \textbf{PAGen}~\cite{acl20-dialog-metrics} are both personalized dialog generation models that employ conditional VAE to exploit textual user information for generating diverse responses. We regard the question and supporting facts as the dialog context and adopt the UGC as the user information to generate personalized answers.
    \item \textbf{Per-TFMR} is developed from \textbf{TFMR} by adding the historical UGC to the supporting facts for incorporating personal information.
    \item \textbf{KOBE}~\cite{kdd19-description} is a personalized product description generation model that considers a variety of factors during generation, including product aspects, user categories, and external knowledge base\footnote{\url{https://github.com/THUDM/KOBE}}. We employ neural topic modeling method, GSM~\cite{icml17-gsm}, to cluster users based on the historical UGC, and regard the supporting facts as the knowledge base.
\end{itemize}

\subsection{Evaluation Metrics}
We adopt ROUGE F1 (R-1, R-2, R-L) and Embedding-based Similarity (ES)~\cite{DBLP:conf/emnlp/LiuLSNCP16}, including
ES-Extreme, ES-Greedy, and ES-Average, as evaluation metrics to measure the performance of answer generation. In addition, following previous studies on personalized text generation~\cite{ijcai19-dialog,acl20-dialog-metrics}, Persona Coverage ($\mathcal{C}_{per}$), Users-Distinct ($uDist$), and User-Language-Perplexity ($uPPL$) are adopted for evaluating the ability of personalization:
\begin{itemize}
    \item Persona Coverage ($\mathcal{C}_{per}$)~\cite{ijcai19-dialog} is adopted to evaluate how well the personal information is reflected in the generated answers at word-level. Suppose that there are $N$ samples with $M$ retrieved personal history snippets for each. The persona coverage score is defined as:
    \begin{equation}
        \mathcal{C}_{per}=\frac{\sum\nolimits^N_i \max_{j\in[1,M]}\mathcal{S}(y_i,p^i_j)}{N},
    \end{equation}
    where $\mathcal{S}(y_i,p^i_j)$ is the mean of the \textit{inverse document frequency} of shared words $W$ between the generated answer $y_i$ and the personal history snippet $p^i_j$:
    \begin{equation}
        \mathcal{S}(y_i,p^i_j)=\frac{\sum\nolimits_{w_k\in W}idf_k}{|W|}.
    \end{equation}
    \item Users-Distinct ($uDist$)~\cite{acl20-dialog-metrics} is adopted to measure the in-group diversity of answers generated within a group of users for the same question.
    \item User-Language-Perplexity ($uPPL$)~\cite{acl20-dialog-metrics} is used to quantify the power of a personalized text generation method on learning the user language style. It adopts a bi-gram language model for perplexity computation, which is pre-trained with the whole training data and then fine-tuned with each user's history content.  
\end{itemize}

\subsection{Implementation Details}
Following the general Transformer settings~\cite{nips17-transformer}, we apply a six-layer Transformer encoder and a two-layer Transformer decoder for all transformer-based models. The embedding size of the input and the hidden size is set to be 512. The size of the Transformer FFN inner representation size is set to be 2048, and ReLU is used as the activation function. The learning rate and the dropout rate are set to be 0.0001 and 0.1, respectively. During training, the batch size is set to be 32, while at the inference phase, we use beam search with a beam size of 10. For each model, we all train for 20 epochs. Based on some preliminary experiments, we find that there is not much difference on the performance when tuning the cluster number within \{5,10,20,40\}. Thus, the cluster number of user preference is set to be 10 for both KOBE and our proposed model. The hyper-parameter $\lambda$ is set to 1 and 0.1 for balancing losses and the regularization term.

\section{Experimental Results}\label{section6}

\subsection{Research Questions}
The empirical analysis targets the following research questions: 
\begin{itemize}
    \item \textbf{RQ1}: How is the performance of the proposed method on answer generation, compared to existing methods? Can answer generation in PQA benefit from personalization?
    \item \textbf{RQ2}: How well does the proposed method generate personalized answers in terms of multi-perspective user preferences? 
    \item \textbf{RQ3}: How does each component in the proposed method contribute to the overall performance?
    \item \textbf{RQ4}: Which aspects of product question answering can be improved by personalization?
\end{itemize}

\subsection{Evaluation on Answer Generation (RQ1)}

\begin{table*}
\centering
\caption{\label{ansgen-elec} Method comparisons of Answer Generation on \textit{Electronics} dataset. $^\dagger$ indicates that the model is better than the best performance of baseline methods (\underline{underline} scores) with statistical significance (measured by significance test at $p<0.05$).}
\begin{tabular}{lcccccc}
\toprule
Model& R-1&R-2&R-L&ES-Ext.&ES-Gre.&ES-Ave.\\
\midrule
BM25&13.2&1.8&11.7&35.4&66.6&84.7 \\
\midrule
S2SAR~\cite{wsdm19-answer-gen-gao} &13.5&2.3&12.5&35.8&65.6&78.9\\
RAGE~\cite{wsdm19-answer-gen-chen} &13.2&2.1&12.0&35.6&65.5&80.2\\
TFMR~\cite{acl19-answer-gen}&14.8&2.5&13.2&36.1&65.1&\underline{85.2}\\
\midrule
Per-S2SAR~\cite{acl18-dialog} &14.0&2.5&13.0&35.6&65.5&82.6\\
Per-CVAE~\cite{ijcai19-dialog} &13.3&2.0&11.9&35.1&63.4&85.0\\
PAGen~\cite{acl20-dialog-metrics} &13.5&2.1&12.0&35.5&64.4&84.6\\
Per-TFMR&\underline{15.1}&\underline{2.7}&\underline{13.5}&36.3&66.9&85.0\\
KOBE~\cite{kdd19-description} &15.0&2.5&13.3&\underline{36.8}&\underline{67.4}&84.9\\
\midrule
\midrule
\textbf{PAGE}&\textbf{16.9}$\dagger$&\textbf{3.5}$\dagger$&\textbf{15.1}$\dagger$&38.3$\dagger$&67.7$\dagger$&85.0\\
- w/o PHI&16.2&3.3&14.5&37.5&67.1&85.1\\
- w/o PPM&16.3&3.5&14.6&\textbf{38.6}&\textbf{67.9}&\textbf{85.3}\\
- w/o PIS&15.7&2.8&13.9&37.1&66.6&85.2\\

\bottomrule
\end{tabular}
\end{table*}

We first summarize the evaluation results on answer generation performance in Table~\ref{ansgen-elec},~\ref{ansgen-home}, and~\ref{ansgen-sport}. In general, our proposed method, PAGE, consistently and substantially outperforms all these strong baselines across three different domains, not only the general answer generation models for PQA, but also the adapted personalized text generation models. Specifically, there are several notable observations as follows:

(1) The retrieved fact snippets obtained by BM25 are far from satisfying as an answer to the given question. 

(2) There is an obvious gap between non-persona models and their persona-aware variants. For instance, Per-S2SAR and Per-TFMR achieve about 1\% improvement by incorporating user information into S2SAR and TFMR. The results validate the effectiveness of taking into account the user information for answer generation in E-Commerce scenario.

(3) The results also show that, compared to RNN-based models (e.g., S2SAR, RAGE, Per-S2SAR, Per-CVAE), Transformer-based models (e.g., TFMR, Per-TFMR, KOBE) improve the natural language generation performance to a great extent.

(4) Among those persona-aware generation baselines, Per-TFMR achieves the best performance on content preservation. However, KOBE has a similar performance with TFMR, which indicates that only using user category feature is inadequate to improve the answer generation for product-related questions, compared to introducing personal history as external knowledge. 

(5) Finally, our proposed method, PAGE, outperforms all the baselines by a noticeable margin across three product categories. The results on answer generation show that PAGE can guarantee the quality of generated answers when exploring the personalization in PQA. 

Overall, this experiment provides the answer to \textbf{RQ1}: \textit{(i) The proposed method substantially and consistently outperforms existing methods on the evaluation of answer generation. (ii) Answer generation in PQA can benefit from certain personalization techniques.}

\begin{table*}
\centering
\caption{\label{ansgen-home} Method comparisons of Answer Generation on \textit{Home\&Kitchen} dataset.  $^\dagger$ indicates that the model is better than the best performance of baseline methods (\underline{underline} scores) with statistical significance (measured by significance test at $p<0.05$).}
\begin{tabular}{lcccccc}
\toprule
Model& R-1&R-2&R-L&ES-Ext.&ES-Gre.&ES-Ave.\\
\midrule
BM25&12.4&2.0&12.1&36.4&66.6&85.7 \\
\midrule
S2SAR~\cite{wsdm19-answer-gen-gao} &12.2&2.0&11.4&37.1&66.6&85.1\\
RAGE~\cite{wsdm19-answer-gen-chen} &13.0&2.4&12.0&37.2&66.8&85.5\\
TFMR~\cite{acl19-answer-gen}&15.5&2.7&13.9&37.2&66.9&85.8\\
\midrule
Per-S2SAR~\cite{acl18-dialog} &14.6&2.7&13.4&37.1&66.4&84.1\\
Per-CVAE~\cite{ijcai19-dialog} &14.8&2.7&13.9&33.9&67.0&84.6\\
PAGen~\cite{acl20-dialog-metrics} &15.3&2.9&14.1&35.9&66.8&85.2\\
Per-TFMR&\underline{16.1}&\underline{3.0}&14.4&\underline{38.0}&\underline{67.1}&\underline{86.2}\\
KOBE~\cite{kdd19-description}& 16.0&2.8&\underline{14.5}&37.0&66.8&85.7\\
\midrule
\midrule
\textbf{PAGE}&17.8$\dagger$&\textbf{3.9}$\dagger$&16.0$\dagger$&\textbf{40.6}$\dagger$&\textbf{68.8}$\dagger$&\textbf{86.5}$\dagger$\\
- w/o PHI&17.5&3.6&15.6&39.5&68.3&86.5\\
- w/o PPM&\textbf{18.0}&3.8&\textbf{16.2}&40.1&68.5&86.3\\
- w/o PIS&16.9&3.3&14.8&38.2&66.9&86.0\\

\bottomrule
\end{tabular}
\end{table*}

\begin{table*}
\centering
\caption{\label{ansgen-sport} Method comparisons of Answer Generation on \textit{Sports\&Outdoors} dataset.  $^\dagger$ indicates that the model is better than the best performance of baseline methods (\underline{underline} scores) with statistical significance (measured by significance test at $p<0.05$).}
\begin{tabular}{lcccccc}
\toprule
Model& R-1&R-2&R-L&ES-Ext.&ES-Gre.&ES-Ave.\\
\midrule
BM25&13.0&1.9&11.8&35.8&66.1&85.6 \\
\midrule
S2SAR~\cite{wsdm19-answer-gen-gao} &13.4&1.9&12.5&37.1&66.6&84.7\\
RAGE~\cite{wsdm19-answer-gen-chen} &13.5&1.9&12.5&35.6&65.9&85.2\\
TFMR~\cite{acl19-answer-gen}&13.9&2.5&12.7&34.8&65.5&85.3\\
\midrule
Per-S2SAR~\cite{acl18-dialog} &13.7&2.3&12.7&36.0&65.6&84.9\\
Per-CVAE~\cite{ijcai19-dialog} &11.9&1.9&10.4&32.9&65.6&84.5\\
PAGen~\cite{acl20-dialog-metrics} &13.9&2.3&12.6&35.6&66.5&84.9\\
Per-TFMR&\underline{14.7}&\underline{3.3}&\underline{13.4}&36.7&\underline{67.1}&85.1\\
KOBE~\cite{kdd19-description} &13.9&2.3&13.0&\underline{36.7}&66.2&\underline{85.7}\\
\midrule
\midrule
\textbf{PAGE}&\textbf{17.4}$\dagger$&\textbf{4.3}$\dagger$&\textbf{15.8}$\dagger$&39.3$\dagger$&68.6$\dagger$&86.4$\dagger$\\
- w/o PHI&16.9&3.1&15.0&39.2&\textbf{68.8}&\textbf{87.0}\\
- w/o PPM&17.0&4.2&15.5&\textbf{39.4}&68.5&86.5\\
- w/o PIS&14.4&3.4&13.8&37.2&67.1&85.8\\

\bottomrule
\end{tabular}
\end{table*}

\begin{table*}
\centering
\caption{\label{persona-gen-elec} Method comparisons of Personalization on \textit{Electronics} dataset. The higher$\uparrow$ (or lower$\downarrow$), the better. We only compare the user perplexity ($uPPL$) among Transformer-based models, since the language probability distribution is different from RNN-based models and ORACLE answers so that their $uPPL$ scores are marked as "-". $^\dagger$ indicates that the model is better than the best performance of baseline methods (\underline{underline} scores) with statistical significance (measured by significance test at $p<0.05$).
}
\begin{tabular}{lcccc}
\toprule
Model&$\mathcal{C}_{per}$$\uparrow$ & $uDist$-1$\uparrow$ & $uDist$-2$\uparrow$ & $uPPL$$\downarrow$\\
\midrule
TFMR~\cite{nips17-transformer}&0.0544&0.6205&0.7194&390.0\\
\midrule
Per-S2SAR~\cite{acl18-dialog} &0.0372&0.3832&0.4967&\small{-}\\
Per-CVAE~\cite{ijcai19-dialog} &0.0248&0.5846&0.7058&-\\
PAGen~\cite{acl20-dialog-metrics} &0.0359&0.5456&0.6674&-\\
Per-TFMR&0.0552&\underline{0.6925}&\underline{0.8583}&\underline{377.5}\\
KOBE~\cite{kdd19-description} &\underline{0.0608}&0.6449&0.7645&381.9\\
\midrule
ORACLE&0.2121&0.7300&0.9481&-\\
\midrule
\midrule
\textbf{PAGE}&\textbf{0.1320}$\dagger$&0.7229$\dagger$&\textbf{0.8990}$\dagger$&337.1$\dagger$\\
- w/o PHI&0.0935&0.7031&0.8507&\textbf{321.2}\\
- w/o PPM&0.1299&0.7003&0.8551&372.8\\
- w/o PIS&0.0690&\textbf{0.7326}&0.8876&361.7\\
\bottomrule
\end{tabular}
\end{table*}

\subsection{Evaluation on Personalization (RQ2)}\label{sec:exp-per}
The evaluation results of personalization on generated answers are given in Table~\ref{persona-gen-elec},~\ref{persona-gen-home}, and~\ref{persona-gen-sport}. It is observed that our proposed method makes a remarkable performance on generating personalized answers according to the evaluation metrics of personalized text generation. Note that \textbf{ORACLE} represents the evaluation results of ground-truth/reference answers.  Although TFMR achieves the best performance on answer generation among non-persona methods, compared to ORACLE, the performance of personalization is far from ideal. 
On the other hand, the ORACLE results on $\mathcal{C}_{per}$ show that the user-generated answers share certain personal information with the historical UGC whether in knowledge-, aspect-, or vocabulary-level, which explicitly reflects the corresponding user preference towards the given question. 
Meanwhile, the ORACLE results on $uDist$ indicate the diversity of answers as well as user preference among different users towards the same question. 

For those RNN-based text generation methods (Per-S2SAR \& Per-CVAE), the repetition issue~\cite{acl17-pointer} leads to relatively low quality in the generated answers. 
Thus, we mainly discuss the performance comparisons among Transformer-based models. 
Per-TFMR and KOBE generally achieve better performance on personalization than TFMR by taking into account user personal information. In specific, their improvement on persona coverage and user perplexity metrics is relatively trivial, while Per-TFMR significantly improves the diversity of the generated answers than KOBE, indicating that individual user modeling can bring more diversity to generated answers than group-based user modeling. As for our proposed model, we observe that it makes substantial improvements on these personalization metrics across different product domains. First, the performance boosting on $\mathcal{C}_{per}$ indicates that PAGE effectively reflects personal information in the generated answers. Second, the results on $uDist$ are very closed to the ORACLE results, which shows the promising diversity of the generated answers by PAGE. Third, the improvement on $uPPL$ shows that PAGE can generate answers more closely with user-preferred language styles. 

This evaluation on personalization-related metrics provides the answer to \textbf{RQ2}: \textit{The proposed method effectively generates personalized answers with  a higher diversity of user-centric information as well as user-preferred language styles. }

\begin{table*}
\centering
\caption{\label{persona-gen-home} Method comparisons of Personalization on \textit{Home\&Kitchen} dataset. 
The higher$\uparrow$ (or lower$\downarrow$), the better. We only compare the user perplexity ($uPPL$) among Transformer-based models, since the language probability distribution is different from RNN-based models and ORACLE answers so that their $uPPL$ scores are marked as "-". $^\dagger$ indicates that the model is better than the best performance of baseline methods (\underline{underline} scores) with statistical significance (measured by significance test at $p<0.05$).}
\begin{tabular}{lcccc}
\toprule
Model&$\mathcal{C}_{per}$$\uparrow$ & $uDist$-1$\uparrow$ & $uDist$-2$\uparrow$ & $uPPL$$\downarrow$\\
\midrule
TFMR~\cite{nips17-transformer}&0.0567&0.5952&0.6827&319.0\\
\midrule
Per-S2SAR~\cite{acl18-dialog} &0.0395&0.3756&0.4862&-\\
Per-CVAE~\cite{ijcai19-dialog} &0.0268&0.5380&0.6696&-\\
PAGen~\cite{acl20-dialog-metrics} &0.0326&0.5008&0.6327&-\\
Per-TFMR&0.0578&\underline{0.6654}&\underline{0.8498}&\underline{312.3}\\
KOBE~\cite{kdd19-description} &\underline{0.0640}&0.6081&0.7096&374.0\\
\midrule
ORACLE&0.2013&0.7170&0.9465&-\\
\midrule
\midrule
\textbf{PAGE}&\textbf{0.1342}$\dagger$&\textbf{0.7027}$\dagger$&\textbf{0.8769}$\dagger$&\textbf{296.1}$\dagger$\\
- w/o PHI&0.1159&0.6409&0.8254&300.3\\
- w/o PPM&0.1327&0.6634&0.8508&315.1\\
- w/o PIS&0.0651&0.6978&0.8675&298.1\\
\bottomrule
\end{tabular}
\end{table*}

\begin{table*}
\centering
\caption{\label{persona-gen-sport} Method comparisons of Personalization on \textit{Sports\&Outdoors} dataset. The higher$\uparrow$ (or lower$\downarrow$), the better. We only compare the user perplexity ($uPPL$) among Transformer-based models, since the language probability distribution is different from RNN-based models and ORACLE answers so that their $uPPL$ scores are marked as "-". $^\dagger$ indicates that the model is better than the best performance of baseline methods (\underline{underline} scores) with statistical significance (measured by significance test at $p<0.05$).
}
\begin{tabular}{lcccc}
\toprule
Model&$\mathcal{C}_{per}$$\uparrow$ & $uDist$-1$\uparrow$ & $uDist$-2$\uparrow$ & $uPPL$$\downarrow$\\
\midrule
TFMR~\cite{nips17-transformer}&0.0374&0.6215&0.7071&331.5\\
\midrule
Per-S2SAR~\cite{acl18-dialog} &0.0213&0.3490&0.4397&-\\
Per-CVAE~\cite{ijcai19-dialog} &0.0172&0.6343&0.6764&-\\
PAGen~\cite{acl20-dialog-metrics} &0.0224&0.5994&0.6412&-\\
Per-TFMR&\underline{0.0391}&\underline{0.7090}&\underline{0.8675}&\underline{310.0}\\
KOBE~\cite{kdd19-description} &0.0347&0.6713&0.7326&334.6\\
\midrule
ORACLE&0.2298&0.7312&0.9519&-\\
\midrule
\midrule
\textbf{PAGE}&\textbf{0.1448}$\dagger$&0.7367$\dagger$&\textbf{0.8839}$\dagger$&284.7$\dagger$\\
- w/o PHI&0.0967&0.7209&0.8697&\textbf{261.8}\\
- w/o PPM&0.1190&0.7062&0.8692&304.2\\
- w/o PIS&0.0778&\textbf{0.7382}&0.8738&306.2\\
\bottomrule
\end{tabular}
\end{table*}

\subsection{Effectiveness of Components (RQ3)}
\subsubsection{\textbf{Ablation Study}}
Since we propose three specific modules, including Persona History Incorporation (PHI), Persona Preference Modeling (PPM), and Personal Information Summarizer (PIS), for modeling personal preference from different perspectives, we conduct ablation study to testify the importance of these modules in answer generation and personalization. In specific, ``w/o PPM" means we only take the retrieved historical UGC  $\mathcal{H^*}$ as the additional personal information for PHI module to capture the knowledge-level user preferences, while the PPM module is non-functional. 
``w/o PHI" means we only take the BoW representation of $\mathcal{H^*}$ as the additional personal information for PPM module to capture the aspect-level and vocabulary-level user preferences, while the PHI module is disabled. 
``w/o PIS" means we take the original vocabulary distribution of the Transformer decoder to generate the answers without the consideration of the user vocabulary and the words in the source text.  

As for the content-preserving metrics in Table~\ref{ansgen-elec},~\ref{ansgen-home}, and~\ref{ansgen-sport}, it can be observed that PIS module contributes to the most significant improvement to the final performance, which is attributed to two main reasons. First, in accordance with the findings in Section~\ref{sec:exp-per}, people are likely to talk with their own knowledge or wording habits when writing answers. 
Second, in fact, PIS module serves as a summarizer to integrate all the learned personal information, rather than a simple component. Comparatively, discarding PHI or PPM modules makes not much difference in the content-preserving performance. We conjecture that these two modules can serve as a complementary component to fetch personal information for the PIS module to generate persona-aware answers (one from knowledge-level, the other from aspect-level and vocabulary-level). 

As for the personalization metrics in Table~\ref{persona-gen-elec},~\ref{persona-gen-home}, and~\ref{persona-gen-sport}, we can find that both PHI and PPM modules contribute to the diversity of the generated answers among different users, e.g., $uDist$. This indicates that the answers to the same question actually vary in terms of user preferences towards different product aspects or practical knowledge from person to person. Separately, PHI module successfully incorporates user history into the generated answers as shown in the $\mathcal{C}_{per}$ metric, while PPM module is useful for generating answers that are more consistent with the user's language style, as reflected by $uPPL$. The results demonstrate the effectiveness of PHI and PPM modules on incorporating personal knowledge and building user-specific vocabulary. Conversely, PIS module makes little contribution to diversifying the generated answers, but it guarantees the model to effectively leverage user historical information and preference-based user vocabulary for answer generation.

Overall, \textit{the multi-perspective preference modeling contributes to the personalization of PQA from different perspectives and aids in generating answers with remarkable content quality}, which is the answer to the \textbf{RQ3}.

\subsubsection{\textbf{Latent Aspect-level Preference Clustering}}

The Persona Preference Modeling module is proposed to extract latent aspect-level preference features from user history for guiding personalized answer generation. To facilitate further investigation of the aspect-level preference modeling, we derive the semantic similarity matrix $\beta$ from Equation~\ref{eq:beta} in each category, and rank the top representative words for each preference cluster. The results of clusters and words for Electronics and Home\&Kitchen categories are presented in Table~\ref{cluster}. Note that the preference type of each cluster is inferred from the cluster results, since there is no ground-truth label for the latent cluster. From the personal preference modeling result, we observe that the PPM module effectively cluster users into different preference categories, with different representative words. 
There are obvious inclinations for each cluster in different product domains.  
In the Electronics category, user-preferred aspects are clustered into different groups, such as \textbf{Camera}, \textbf{Sound}, and \textbf{Device Configuration}, etc. 
For instance, some users may prefer to know about the device configuration of certain electronic products, while some may care more about the photography related information. 
In the Home\&Kitchen category, it can be observed that there are some users who prefer to \textbf{Drink\&Food}, \textbf{Cooker\&Material}, or \textbf{Size\&Quality}. Meanwhile, there are some preference clusters shared across different categories, e.g. \textbf{Price}. 
The clustering results answer \textbf{RQ3} from a different angle:  \textit{On one hand, personal preference modeling can provide aspect-level user preference information as well as assist in building a preference-aware user vocabulary, which contributes to the personalization of answer generation. On the other hand, such latent topic modeling approach provides an unsupervised way to discover the latent features of different users based on their history content. }

\begin{table*}
\fontsize{9}{10}\selectfont
    \centering
    \caption{Top representative words for each aspect-level preference cluster on two different product categories}
    \begin{tabular}{p{1.6cm}<{\centering}|p{2.6cm}<{\centering}|p{2.6cm}<{\centering}|p{2.6cm}<{\centering}|p{2.6cm}<{\centering}|c}
    \toprule
        \textbf{Category} & \textbf{Preference 1} & \textbf{Preference 2}& \textbf{Preference 3}& \textbf{Preference 4}&  $\dots$\\
        \midrule
        & \textbf{Camera} &\textbf{Sound}&\textbf{Configuration}&\textbf{Price}&\multirow{2}{*}{$\dots$}\\
        \cmidrule(lr){2-5}
        \textbf{Electronics}& Camera, Lens, Canon, Nikon, Flash, Zoom, Sony, Focus, Pictures & Sound, TV, Quality, Cable, Speakers, Radio, Audio, Headphones &USB, Drive, Port, Windows, Device, Computer, Software, Player, HDMI&Product, Price, Nice, Bought, Buy, Recommend, Bit, Purchase, Money&\\
        \midrule
         & \textbf{Drink\&Food}& \textbf{Cooker\&Material}& \textbf{Size\&Quality}& \textbf{Price}&\multirow{2}{*}{$\dots$}\\
        \cmidrule(lr){2-5}
        \textbf{Home\& Kitchen}& Coffee, Water, Pot, Cup, Tea, Easy, Hot, Food, Bread, Rice, Milk, Taste   & Pan, Machine, Oven, Cooker, Maker, Stainless, Plastic, Steel, Cooking &  Size, Fit, Nice, Quality, Perfect, Comfortable, Sturdy, Hold, Heavy & Unit, Product, Bought, Buy, Price, Worth, Purchase, Recommend, Money&\\
        \bottomrule
    \end{tabular}
    \label{cluster}
    \vspace{-0.2cm}
\end{table*}

\subsection{Discussions (RQ4)}
\subsubsection{\textbf{Human Evaluation}}
To further evaluate the quality of generated answers from each model, we also conduct human evaluation to evaluate the generated answers by scoring the following six aspects: 
\begin{itemize}
    \item \textit{Informativeness}: how rich is the generated answer in information towards the given question? 
    \item \textit{Fluency}: how fluent is the generated answer?
    \item \textit{Relevancy}: : how relevant is the generated answer to the given question?
    \item \textit{Diversity}: how diverse are the generated answers for different users towards the same question?
    \item \textit{Persona}: how well does the generated answer reflect the user's personal preference?
    \item \textit{Overall Quality}: : how helpful is the generated answer to the user according to the overall quality?
\end{itemize}
We randomly sample 50 instances from each category and generate their answers with four methods, including TFMR, Per-TFMR, KOBE, and the proposed PAGE. Three annotators are recruited to score each generated answer with 1 to 5 (higher the better). 
These annotators are all well-educated research assistants with a related background of NLP. 
However, since the degree of \textit{Persona} is difficult to be quantified by human judgment, we simplify the scoring into a binary score, representing the judgment of whether the generated answer can reflect the user's persona. 

Table~\ref{human_eval} summarizes the results of human evaluation. It can be observed that our model, PAGE, consistently outperforms these strong baselines from different aspects of human evaluation. It is noteworthy that the \textit{Informativeness}, \textit{Diversity}, and \textit{Persona} are significantly improved by PAGE, which demonstrates that PAGE effectively enriches and diversifies the generated answers by taking into account user preference. According to the scores of \textit{Overall Quality}, which measures the quality of the generated answers against the reference answers, it is beneficial to exploit personalization for product question answering in E-Commerce. 

The human judgements provide the answer to the \textbf{RQ4}: \textit{The answers generated by the proposed method preserve a higher degree of informativeness, diversity as well as explicitly reflect some user-centric persona, which contributes to a higher overall quality. }

\begin{table}
\caption{Human evaluation results. The Fleiss’ kappa of the annotations is 0.69, which indicates ``substantial agreement”, and the final scores are calculated by average.}
\centering
  \begin{tabular}{lcccccc}
  \toprule
  Method & Informativeness & Fluency & Relevancy & Diversity & Persona & Overall Quality\\
  \midrule
  TFMR &2.78&3.93&3.42&1.00&1.7\%&1.90\\
  Per-TFMR &3.02&3.97&3.53&3.15&7.2\%&2.17\\
  KOBE &2.85&3.68&3.18&1.75&5.0\%&1.98\\
  \textbf{PAGE} & \textbf{3.30}&\textbf{4.02}&\textbf{3.57}&\textbf{3.70}&\textbf{15.8\%}&\textbf{2.78}\\
  \bottomrule
  \end{tabular}

\label{human_eval}
\end{table}

\subsubsection{\textbf{Case Study}}

\begin{table}
\caption{Case study}\label{case}
\centering
  \begin{tabular}{cp{11cm}}
  \toprule
  \textbf{Question}& While a TV is mounted, can you also mount a sound bar underneath the TV? \\
  \midrule
  \multirow{3}{*}{\textbf{\shortstack{Supporting\\Fact\\Snippets\\(Partial)}}}& 1. Whats also nice is that I have a \textbf{sound bar}, so I just simply placed it on the top shelf of the stand and it sounds as if the \textbf{audio} is coming right through the TV! \\
  & 2. It is sturdy, holds a \textbf{50 inch TV} and swivels with no problems. \\
  & 3. I have my Viera \textbf{55''}; mounted on this and it is perfect! \\
  \midrule
  \midrule
  \textbf{UGC (User1)}&  I 've used these to mount a pair of \textbf{Polk Audio RTI A1 speakers}, ... // ... so they most definantly will hold a 12 pound \textbf{speaker}. // Peerless is known in the \textbf{audio} world for making quality products.\\
  \textbf{Reference1}&  Yes . The TV can be mounted on a pedestal that is attached to the back of the stand . This allows you to use the top shelf for a soundbar or center channel \textbf{speaker}. \\
  \midrule
  \textbf{TFMR}&  Yes it can be mounted on the wall.\\
  \textbf{Per-TFMR}&  Yes, the TV can be mounted on either the left or right side.\\
  \textbf{KOBE}&  Yes, the TV can be removed from the floor. The stand is designed so that the stand will stand up against the wall.\\
  \textbf{PAGE}&  Yes. The TV can be mounted on a pedestal that is attached to the top of a cabinet. It is designed to sit on a horizontal TV \textbf{speaker} and secured into the TV.\\
  \midrule
  \midrule
  \textbf{UGC (User2)}&  This TV stand is \textbf{perfect} for \textbf{our new 50'' SONY TV} that we bought from Amazon at the same time.\\
  \textbf{Reference2}&  With \textbf{our 50'' Sony} mounted on the lowest position on the mounting bars, I have 6-3/8'' from the top glass shelf to the bottom of the TV frame. If you move the TV up to the top of the mounting bars you would have about 12'' of space. \\
  \midrule
  \textbf{TFMR}&  Yes it can be mounted on the wall.\\
  \textbf{Per-TFMR}&  Yes. The TV can be mounted on either side or left or right or left. There is a mounting bracket on the back of the TV.\\
  \textbf{KOBE}&  Yes, the TV can be adjusted vertically and the sides of the TV will stand up from the floor. It is a great value for the price.\\
  \textbf{PAGE}&  Yes. We have a \textbf{50 inch TV} (slides left to right) and it fits \textbf{perfectly}.\\
  \midrule
  \midrule
  \textbf{UGC (User3)} &  Quality of the unit was excellent and sturdy. \textbf{I mounted my 55'' LED Sony TV to the stand.} ... // \textbf{I don't see why} it couldn't be used for other bake goods. ... \\
  \textbf{Reference3}&  \textbf{The TV I have is a 55 inch.} I use a stereo for surround sound, but I have a 5 inch center speaker under the TV. \\
  \midrule
  \textbf{TFMR}&  Yes it can be mounted on the wall.\\
  \textbf{Per-TFMR}&  It can be mounted on the back of the TV. The stand is on the top of the TV. There is no way to mount it to the wall.\\
  \textbf{KOBE}&  Yes, the TV can be removed from the floor. The stand is designed so that the stand will stand up against the wall.\\
  \textbf{PAGE}&  \textbf{I don't see why} not. I have a \textbf{55 inch TV.} I put it on the TV and it works great.\\
  \bottomrule
  \end{tabular}
\end{table}

In order to intuitively differentiate our proposed method from other baselines, Table~\ref{case} presents a specific product-related question from Amazon QA. Even though the question is asking for a YES or NO answer, different former buyers still provide diverse responses (\textbf{Reference}) reflecting their own preferences. 
Besides, it can also be observed that the user-generated answers (\textbf{Reference}) typically share some similar preferences with their historical \textbf{UGC}. For instance, \textbf{User1} has a strong preference towards the product aspect of sound quality, which is reflected in both his/her historical posted content and his/her answer to the given question, while \textbf{User2} and \textbf{User3} tend to share their knowledge from previous shopping experiences, which have already been shared in reviews before. 

For the baseline methods, non-persona methods (e.g., \textbf{TFMR}) generate a unified answer for all individual users, which only involves the basic information obtained from the common opinion of supporting facts. Users' personal information enables personalized text generation methods (e.g., \textbf{Per-TFMR} and \textbf{KOBE}) to provide answers with additional information for different users accordingly. 
Among them, the answers generated by \textbf{Per-TFMR} are more diverse than those generated by \textbf{KOBE}. \textbf{KOBE} only captures group-based preference, as it models user features by classifying users into several pre-defined categories, while \textbf{Per-TFMR} employs historical UGC as textual personal information for individual user modeling. 
However, both of them fail to explicitly incorporate user preference into the generated answers. That is to say, although they improve the diversity of the generated answers for different users, the personalization is not actually reflected in the answers. 

With multi-perspective preference modeling, \textbf{PAGE} effectively generates informative and customized answers involving specific user preferences. Similar to the reference answers, \textbf{PAGE} provides personalized answers for \textbf{User1} reflecting his/her aspect-level preference towards sound equipment. As for \textbf{User2} and \textbf{User3}, the generated answers are combined with the practical knowledge from supporting facts for satisfying their individual information needs. Besides, we notice that \textbf{PAGE} can capture some vocabulary-level preference or preferred language style, e.g., \textbf{User2} likes to use ``perfect" to comment on the product, while \textbf{User3} prefer to start talking with ``I don't see why". Overall, the case study provides a piece of intuitive evidence to the \textbf{RQ4}.

\section{Conclusions and Future Work}\label{section7}
In this paper, we investigate the problem of personalized answer generation in E-Commerce, which aims at exploring personal information from historical user-generated content to aid in generating personalized answers to product-related questions. 
Specifically, we propose a novel personalized answer generation method with multi-perspective preference modeling to comprehensively capture user preference for generating persona-aware answers. 
A persona-aware pointer-generator network, which bridges neural topic modeling and pointer network, is developed to dynamically learn the user preference and the user-specific vocabulary for generating persona-aware answers. 
The proposed method outperforms existing methods on real-world E-Commerce QA datasets and effectively generates informative and customized answers for different users by incorporating various user preferences from the perspective of knowledge-, aspect-, and vocabulary-level. In addition, the empirical analysis shows that answer generation in E-Commerce can benefit from personalization.

This work is the first attempt towards personalized answer generation for PQA. There are some limitations and room for further improvement. For example, besides historical UGC, there are other types of user data for user modeling in E-Commerce, such as user-item interaction. It can be beneficial to leverage such heterogeneous or multi-modal data for better user preference modeling. Similar to recent personalized justification or explanation generation studies for item recommendation, another future direction can be to develop graph-based reasoning methods for building explainable PQA systems with personalization. 
Besides, the personalized answer generation study can be extended to other personalized text generation applications in E-Commerce, or other user-centric QA forums, such as Stack Overflow, etc.


\bibliographystyle{ACM-Reference-Format}

\bibliography{sample-base}


\begin{thebibliography}{72}


\ifx \showCODEN    \undefined \def \showCODEN     #1{\unskip}     \fi
\ifx \showDOI      \undefined \def \showDOI       #1{#1}\fi
\ifx \showISBNx    \undefined \def \showISBNx     #1{\unskip}     \fi
\ifx \showISBNxiii \undefined \def \showISBNxiii  #1{\unskip}     \fi
\ifx \showISSN     \undefined \def \showISSN      #1{\unskip}     \fi
\ifx \showLCCN     \undefined \def \showLCCN      #1{\unskip}     \fi
\ifx \shownote     \undefined \def \shownote      #1{#1}          \fi
\ifx \showarticletitle \undefined \def \showarticletitle #1{#1}   \fi
\ifx \showURL      \undefined \def \showURL       {\relax}        \fi
\providecommand\bibfield[2]{#2}
\providecommand\bibinfo[2]{#2}
\providecommand\natexlab[1]{#1}
\providecommand\showeprint[2][]{arXiv:#2}

\bibitem[\protect\citeauthoryear{Ai, Hill, Vishwanathan, and Croft}{Ai
  et~al\mbox{.}}{2019}]%
        {cikm19-product-search}
\bibfield{author}{\bibinfo{person}{Qingyao Ai}, \bibinfo{person}{Daniel~N.
  Hill}, \bibinfo{person}{S.~V.~N. Vishwanathan}, {and}
  \bibinfo{person}{W.~Bruce Croft}.} \bibinfo{year}{2019}\natexlab{}.
\newblock \showarticletitle{A Zero Attention Model for Personalized Product
  Search}. In \bibinfo{booktitle}{\emph{CIKM}}. \bibinfo{pages}{379--388}.
\newblock


\bibitem[\protect\citeauthoryear{Ai, Zhang, Bi, Chen, and Croft}{Ai
  et~al\mbox{.}}{2017}]%
        {sigir17-product-search}
\bibfield{author}{\bibinfo{person}{Qingyao Ai}, \bibinfo{person}{Yongfeng
  Zhang}, \bibinfo{person}{Keping Bi}, \bibinfo{person}{Xu Chen}, {and}
  \bibinfo{person}{W.~Bruce Croft}.} \bibinfo{year}{2017}\natexlab{}.
\newblock \showarticletitle{Learning a Hierarchical Embedding Model for
  Personalized Product Search}. In \bibinfo{booktitle}{\emph{SIGIR}}.
  \bibinfo{pages}{645--654}.
\newblock


\bibitem[\protect\citeauthoryear{Bauer, Wang, and Bansal}{Bauer
  et~al\mbox{.}}{2018}]%
        {emnlp18-answer-gen1}
\bibfield{author}{\bibinfo{person}{Lisa Bauer}, \bibinfo{person}{Yicheng Wang},
  {and} \bibinfo{person}{Mohit Bansal}.} \bibinfo{year}{2018}\natexlab{}.
\newblock \showarticletitle{Commonsense for Generative Multi-Hop Question
  Answering Tasks}. In \bibinfo{booktitle}{\emph{EMNLP}}.
  \bibinfo{pages}{4220--4230}.
\newblock


\bibitem[\protect\citeauthoryear{Bi, Wu, Yan, Wang, Xia, and Li}{Bi
  et~al\mbox{.}}{2019}]%
        {emnlp19-answer-gen}
\bibfield{author}{\bibinfo{person}{Bin Bi}, \bibinfo{person}{Chen Wu},
  \bibinfo{person}{Ming Yan}, \bibinfo{person}{Wei Wang},
  \bibinfo{person}{Jiangnan Xia}, {and} \bibinfo{person}{Chenliang Li}.}
  \bibinfo{year}{2019}\natexlab{}.
\newblock \showarticletitle{Incorporating External Knowledge into Machine
  Reading for Generative Question Answering}. In
  \bibinfo{booktitle}{\emph{EMNLP-IJCNLP}}. \bibinfo{pages}{2521--2530}.
\newblock


\bibitem[\protect\citeauthoryear{Bi, Ai, and Croft}{Bi et~al\mbox{.}}{2020}]%
        {review4productsearch}
\bibfield{author}{\bibinfo{person}{Keping Bi}, \bibinfo{person}{Qingyao Ai},
  {and} \bibinfo{person}{W.~Bruce Croft}.} \bibinfo{year}{2020}\natexlab{}.
\newblock \showarticletitle{A Review-based Transformer Model for Personalized
  Product Search}.
\newblock \bibinfo{journal}{\emph{CoRR}}  \bibinfo{volume}{abs/2004.09424}
  (\bibinfo{year}{2020}).
\newblock
\urldef\tempurl%
\url{https://arxiv.org/abs/2004.09424}
\showURL{%
\tempurl}


\bibitem[\protect\citeauthoryear{Carmel, Lewin{-}Eytan, and Maarek}{Carmel
  et~al\mbox{.}}{2018}]%
        {sigir18-pqa-challenge}
\bibfield{author}{\bibinfo{person}{David Carmel}, \bibinfo{person}{Liane
  Lewin{-}Eytan}, {and} \bibinfo{person}{Yoelle Maarek}.}
  \bibinfo{year}{2018}\natexlab{}.
\newblock \showarticletitle{Product Question Answering Using Customer Generated
  Content - Research Challenges}. In \bibinfo{booktitle}{\emph{SIGIR}}.
  \bibinfo{pages}{1349--1350}.
\newblock


\bibitem[\protect\citeauthoryear{Chen, Qiu, Yang, Zhou, Huang, Li, and
  Bao}{Chen et~al\mbox{.}}{2019d}]%
        {www19-review-help2}
\bibfield{author}{\bibinfo{person}{Cen Chen}, \bibinfo{person}{Minghui Qiu},
  \bibinfo{person}{Yinfei Yang}, \bibinfo{person}{Jun Zhou},
  \bibinfo{person}{Jun Huang}, \bibinfo{person}{Xiaolong Li}, {and}
  \bibinfo{person}{Forrest~Sheng Bao}.} \bibinfo{year}{2019}\natexlab{d}.
\newblock \showarticletitle{Multi-Domain Gated {CNN} for Review Helpfulness
  Prediction}. In \bibinfo{booktitle}{\emph{WWW}}. \bibinfo{pages}{2630--2636}.
\newblock


\bibitem[\protect\citeauthoryear{Chen, Guan, Zhao, Zhao, Wang, Zhao, and
  Sun}{Chen et~al\mbox{.}}{2019a}]%
        {aaai19-answer-sel}
\bibfield{author}{\bibinfo{person}{Long Chen}, \bibinfo{person}{Ziyu Guan},
  \bibinfo{person}{Wei Zhao}, \bibinfo{person}{Wanqing Zhao},
  \bibinfo{person}{Xiaopeng Wang}, \bibinfo{person}{Zhou Zhao}, {and}
  \bibinfo{person}{Huan Sun}.} \bibinfo{year}{2019}\natexlab{a}.
\newblock \showarticletitle{Answer Identification from Product Reviews for User
  Questions by Multi-Task Attentive Networks}. In
  \bibinfo{booktitle}{\emph{AAAI}}. \bibinfo{pages}{45--52}.
\newblock


\bibitem[\protect\citeauthoryear{Chen, Lin, Zhang, Yang, Zhou, and Tang}{Chen
  et~al\mbox{.}}{2019c}]%
        {kdd19-description}
\bibfield{author}{\bibinfo{person}{Qibin Chen}, \bibinfo{person}{Junyang Lin},
  \bibinfo{person}{Yichang Zhang}, \bibinfo{person}{Hongxia Yang},
  \bibinfo{person}{Jingren Zhou}, {and} \bibinfo{person}{Jie Tang}.}
  \bibinfo{year}{2019}\natexlab{c}.
\newblock \showarticletitle{Towards Knowledge-Based Personalized Product
  Description Generation in E-commerce}. In \bibinfo{booktitle}{\emph{KDD}}.
  \bibinfo{pages}{3040--3050}.
\newblock


\bibitem[\protect\citeauthoryear{Chen, Li, Ji, Zhou, and Chen}{Chen
  et~al\mbox{.}}{2019b}]%
        {wsdm19-answer-gen-chen}
\bibfield{author}{\bibinfo{person}{Shiqian Chen}, \bibinfo{person}{Chenliang
  Li}, \bibinfo{person}{Feng Ji}, \bibinfo{person}{Wei Zhou}, {and}
  \bibinfo{person}{Haiqing Chen}.} \bibinfo{year}{2019}\natexlab{b}.
\newblock \showarticletitle{Review-Driven Answer Generation for Product-Related
  Questions in E-Commerce}. In \bibinfo{booktitle}{\emph{WSDM}}.
  \bibinfo{pages}{411--419}.
\newblock


\bibitem[\protect\citeauthoryear{Cheng, Chang, Zhu, {Catherine Kanjirathinkal},
  and Kankanhalli}{Cheng et~al\mbox{.}}{2019}]%
        {tois19-review4rec-explain}
\bibfield{author}{\bibinfo{person}{Zhiyong Cheng}, \bibinfo{person}{Xiaojun
  Chang}, \bibinfo{person}{Lei Zhu}, \bibinfo{person}{Rose {Catherine
  Kanjirathinkal}}, {and} \bibinfo{person}{Mohan~S. Kankanhalli}.}
  \bibinfo{year}{2019}\natexlab{}.
\newblock \showarticletitle{{MMALFM:} Explainable Recommendation by Leveraging
  Reviews and Images}.
\newblock \bibinfo{journal}{\emph{{ACM} Trans. Inf. Syst.}}
  \bibinfo{volume}{37}, \bibinfo{number}{2} (\bibinfo{year}{2019}),
  \bibinfo{pages}{16:1--16:28}.
\newblock


\bibitem[\protect\citeauthoryear{Deng, Lam, Xie, Chen, Li, Yang, and Shen}{Deng
  et~al\mbox{.}}{2020a}]%
        {wikihowqa}
\bibfield{author}{\bibinfo{person}{Yang Deng}, \bibinfo{person}{Wai Lam},
  \bibinfo{person}{Yuexiang Xie}, \bibinfo{person}{Daoyuan Chen},
  \bibinfo{person}{Yaliang Li}, \bibinfo{person}{Min Yang}, {and}
  \bibinfo{person}{Ying Shen}.} \bibinfo{year}{2020}\natexlab{a}.
\newblock \showarticletitle{Joint Learning of Answer Selection and Answer
  Summary Generation in Community Question Answering}. In
  \bibinfo{booktitle}{\emph{AAAI}}. \bibinfo{pages}{7651--7658}.
\newblock


\bibitem[\protect\citeauthoryear{Deng, Xie, Li, Yang, Lam, and Shen}{Deng
  et~al\mbox{.}}{2022}]%
        {tois-dy}
\bibfield{author}{\bibinfo{person}{Yang Deng}, \bibinfo{person}{Yuexiang Xie},
  \bibinfo{person}{Yaliang Li}, \bibinfo{person}{Min Yang},
  \bibinfo{person}{Wai Lam}, {and} \bibinfo{person}{Ying Shen}.}
  \bibinfo{year}{2022}\natexlab{}.
\newblock \showarticletitle{Contextualized Knowledge-aware Attentive Neural
  Network: Enhancing Answer Selection with Knowledge}.
\newblock \bibinfo{journal}{\emph{{ACM} Trans. Inf. Syst.}}
  (\bibinfo{year}{2022}).
\newblock


\bibitem[\protect\citeauthoryear{Deng, Zhang, and Lam}{Deng
  et~al\mbox{.}}{2020b}]%
        {msg}
\bibfield{author}{\bibinfo{person}{Yang Deng}, \bibinfo{person}{Wenxuan Zhang},
  {and} \bibinfo{person}{Wai Lam}.} \bibinfo{year}{2020}\natexlab{b}.
\newblock \showarticletitle{Multi-hop Inference for Question-driven
  Summarization}. In \bibinfo{booktitle}{\emph{EMNLP}}.
  \bibinfo{pages}{6734--6744}.
\newblock


\bibitem[\protect\citeauthoryear{Deng, Zhang, and Lam}{Deng
  et~al\mbox{.}}{2020c}]%
        {oaag}
\bibfield{author}{\bibinfo{person}{Yang Deng}, \bibinfo{person}{Wenxuan Zhang},
  {and} \bibinfo{person}{Wai Lam}.} \bibinfo{year}{2020}\natexlab{c}.
\newblock \showarticletitle{Opinion-aware Answer Generation for Review-driven
  Question Answering in E-Commerce}. In \bibinfo{booktitle}{\emph{CIKM}}.
  \bibinfo{pages}{255--264}.
\newblock


\bibitem[\protect\citeauthoryear{Deng, Zhang, Li, Yang, Lam, and Shen}{Deng
  et~al\mbox{.}}{2020d}]%
        {sigir20-dy}
\bibfield{author}{\bibinfo{person}{Yang Deng}, \bibinfo{person}{Wenxuan Zhang},
  \bibinfo{person}{Yaliang Li}, \bibinfo{person}{Min Yang},
  \bibinfo{person}{Wai Lam}, {and} \bibinfo{person}{Ying Shen}.}
  \bibinfo{year}{2020}\natexlab{d}.
\newblock \showarticletitle{Bridging Hierarchical and Sequential Context
  Modeling for Question-driven Extractive Answer Summarization}. In
  \bibinfo{booktitle}{\emph{SIGIR}}. \bibinfo{pages}{1693--1696}.
\newblock


\bibitem[\protect\citeauthoryear{Elad, Guy, Novgorodov, Kimelfeld, and
  Radinsky}{Elad et~al\mbox{.}}{2019}]%
        {cikm19-description}
\bibfield{author}{\bibinfo{person}{Guy Elad}, \bibinfo{person}{Ido Guy},
  \bibinfo{person}{Slava Novgorodov}, \bibinfo{person}{Benny Kimelfeld}, {and}
  \bibinfo{person}{Kira Radinsky}.} \bibinfo{year}{2019}\natexlab{}.
\newblock \showarticletitle{Learning to Generate Personalized Product
  Descriptions}. In \bibinfo{booktitle}{\emph{CIKM}}.
  \bibinfo{pages}{389--398}.
\newblock


\bibitem[\protect\citeauthoryear{Fan, Jernite, Perez, Grangier, Weston, and
  Auli}{Fan et~al\mbox{.}}{2019b}]%
        {eli5}
\bibfield{author}{\bibinfo{person}{Angela Fan}, \bibinfo{person}{Yacine
  Jernite}, \bibinfo{person}{Ethan Perez}, \bibinfo{person}{David Grangier},
  \bibinfo{person}{Jason Weston}, {and} \bibinfo{person}{Michael Auli}.}
  \bibinfo{year}{2019}\natexlab{b}.
\newblock \showarticletitle{{ELI5:} Long Form Question Answering}. In
  \bibinfo{booktitle}{\emph{ACL}}. \bibinfo{pages}{3558--3567}.
\newblock


\bibitem[\protect\citeauthoryear{Fan, Feng, Guo, Sun, and Li}{Fan
  et~al\mbox{.}}{2019a}]%
        {www19-review-help1}
\bibfield{author}{\bibinfo{person}{Miao Fan}, \bibinfo{person}{Chao Feng},
  \bibinfo{person}{Lin Guo}, \bibinfo{person}{Mingming Sun}, {and}
  \bibinfo{person}{Ping Li}.} \bibinfo{year}{2019}\natexlab{a}.
\newblock \showarticletitle{Product-Aware Helpfulness Prediction of Online
  Reviews}. In \bibinfo{booktitle}{\emph{WWW}}. \bibinfo{pages}{2715--2721}.
\newblock


\bibitem[\protect\citeauthoryear{Feng, Ren, Zhao, Sun, and Li}{Feng
  et~al\mbox{.}}{2021}]%
        {sigir21-pqa}
\bibfield{author}{\bibinfo{person}{Yue Feng}, \bibinfo{person}{Zhaochun Ren},
  \bibinfo{person}{Weijie Zhao}, \bibinfo{person}{Mingming Sun}, {and}
  \bibinfo{person}{Ping Li}.} \bibinfo{year}{2021}\natexlab{}.
\newblock \showarticletitle{Multi-Type Textual Reasoning for Product-Aware
  Answer Generation}. In \bibinfo{booktitle}{\emph{{SIGIR}}}.
  \bibinfo{pages}{1135--1145}.
\newblock


\bibitem[\protect\citeauthoryear{Gao, Lin, Wang, Wang, Yang, He, and Chu}{Gao
  et~al\mbox{.}}{2020}]%
        {cikm20-review4rec}
\bibfield{author}{\bibinfo{person}{Jingyue Gao}, \bibinfo{person}{Yang Lin},
  \bibinfo{person}{Yasha Wang}, \bibinfo{person}{Xiting Wang},
  \bibinfo{person}{Zhao Yang}, \bibinfo{person}{Yuanduo He}, {and}
  \bibinfo{person}{Xu Chu}.} \bibinfo{year}{2020}\natexlab{}.
\newblock \showarticletitle{Set-Sequence-Graph: {A} Multi-View Approach Towards
  Exploiting Reviews for Recommendation}. In \bibinfo{booktitle}{\emph{CIKM}}.
  \bibinfo{pages}{395--404}.
\newblock


\bibitem[\protect\citeauthoryear{Gao, Chen, Ren, Zhao, and Yan}{Gao
  et~al\mbox{.}}{2021}]%
        {tois21}
\bibfield{author}{\bibinfo{person}{Shen Gao}, \bibinfo{person}{Xiuying Chen},
  \bibinfo{person}{Zhaochun Ren}, \bibinfo{person}{Dongyan Zhao}, {and}
  \bibinfo{person}{Rui Yan}.} \bibinfo{year}{2021}\natexlab{}.
\newblock \showarticletitle{Meaningful Answer Generation of E-Commerce
  Question-Answering}.
\newblock \bibinfo{journal}{\emph{{ACM} Trans. Inf. Syst.}}
  (\bibinfo{year}{2021}).
\newblock


\bibitem[\protect\citeauthoryear{Gao, Ren, Zhao, Zhao, Yin, and Yan}{Gao
  et~al\mbox{.}}{2019}]%
        {wsdm19-answer-gen-gao}
\bibfield{author}{\bibinfo{person}{Shen Gao}, \bibinfo{person}{Zhaochun Ren},
  \bibinfo{person}{Yihong Zhao}, \bibinfo{person}{Dongyan Zhao},
  \bibinfo{person}{Dawei Yin}, {and} \bibinfo{person}{Rui Yan}.}
  \bibinfo{year}{2019}\natexlab{}.
\newblock \showarticletitle{Product-Aware Answer Generation in E-Commerce
  Question-Answering}. In \bibinfo{booktitle}{\emph{WSDM}}.
  \bibinfo{pages}{429--437}.
\newblock


\bibitem[\protect\citeauthoryear{Gu, Ding, Wang, and Yin}{Gu
  et~al\mbox{.}}{2020}]%
        {wsdm20-recsys}
\bibfield{author}{\bibinfo{person}{Yulong Gu}, \bibinfo{person}{Zhuoye Ding},
  \bibinfo{person}{Shuaiqiang Wang}, {and} \bibinfo{person}{Dawei Yin}.}
  \bibinfo{year}{2020}\natexlab{}.
\newblock \showarticletitle{Hierarchical User Profiling for E-commerce
  Recommender Systems}. In \bibinfo{booktitle}{\emph{WSDM}}.
  \bibinfo{pages}{223--231}.
\newblock


\bibitem[\protect\citeauthoryear{Guan, Cheng, He, Zhang, Zhu, Peng, and
  Chua}{Guan et~al\mbox{.}}{2019}]%
        {tois19-review4rec}
\bibfield{author}{\bibinfo{person}{Xinyu Guan}, \bibinfo{person}{Zhiyong
  Cheng}, \bibinfo{person}{Xiangnan He}, \bibinfo{person}{Yongfeng Zhang},
  \bibinfo{person}{Zhibo Zhu}, \bibinfo{person}{Qinke Peng}, {and}
  \bibinfo{person}{Tat{-}Seng Chua}.} \bibinfo{year}{2019}\natexlab{}.
\newblock \showarticletitle{Attentive Aspect Modeling for Review-Aware
  Recommendation}.
\newblock \bibinfo{journal}{\emph{{ACM} Trans. Inf. Syst.}}
  \bibinfo{volume}{37}, \bibinfo{number}{3} (\bibinfo{year}{2019}),
  \bibinfo{pages}{28:1--28:27}.
\newblock


\bibitem[\protect\citeauthoryear{Guo, Cheng, Nie, Wang, Ma, and
  Kankanhalli}{Guo et~al\mbox{.}}{2019}]%
        {tois19-product-search}
\bibfield{author}{\bibinfo{person}{Yangyang Guo}, \bibinfo{person}{Zhiyong
  Cheng}, \bibinfo{person}{Liqiang Nie}, \bibinfo{person}{Yinglong Wang},
  \bibinfo{person}{Jun Ma}, {and} \bibinfo{person}{Mohan~S. Kankanhalli}.}
  \bibinfo{year}{2019}\natexlab{}.
\newblock \showarticletitle{Attentive Long Short-Term Preference Modeling for
  Personalized Product Search}.
\newblock \bibinfo{journal}{\emph{{ACM} Trans. Inf. Syst.}}
  \bibinfo{volume}{37}, \bibinfo{number}{2} (\bibinfo{year}{2019}),
  \bibinfo{pages}{19:1--19:27}.
\newblock


\bibitem[\protect\citeauthoryear{Gupta, Kulkarni, Chanda, Rayasam, and
  Lipton}{Gupta et~al\mbox{.}}{2019}]%
        {ijcai19-amazonqa}
\bibfield{author}{\bibinfo{person}{Mansi Gupta}, \bibinfo{person}{Nitish
  Kulkarni}, \bibinfo{person}{Raghuveer Chanda}, \bibinfo{person}{Anirudha
  Rayasam}, {and} \bibinfo{person}{Zachary~C. Lipton}.}
  \bibinfo{year}{2019}\natexlab{}.
\newblock \showarticletitle{AmazonQA: {A} Review-Based Question Answering
  Task}. In \bibinfo{booktitle}{\emph{IJCAI}}. \bibinfo{pages}{4996--5002}.
\newblock


\bibitem[\protect\citeauthoryear{Indurthi, Yu, Back, and
  Cuay{\'{a}}huitl}{Indurthi et~al\mbox{.}}{2018}]%
        {emnlp18-answer-gen2}
\bibfield{author}{\bibinfo{person}{Sathish~Reddy Indurthi},
  \bibinfo{person}{Seunghak Yu}, \bibinfo{person}{Seohyun Back}, {and}
  \bibinfo{person}{Heriberto Cuay{\'{a}}huitl}.}
  \bibinfo{year}{2018}\natexlab{}.
\newblock \showarticletitle{Cut to the Chase: {A} Context Zoom-in Network for
  Reading Comprehension}. In \bibinfo{booktitle}{\emph{EMNLP}}.
  \bibinfo{pages}{570--575}.
\newblock


\bibitem[\protect\citeauthoryear{Ishida, Torisawa, Oh, Iida, Kruengkrai, and
  Kloetzer}{Ishida et~al\mbox{.}}{2018}]%
        {why-question}
\bibfield{author}{\bibinfo{person}{Ryo Ishida}, \bibinfo{person}{Kentaro
  Torisawa}, \bibinfo{person}{Jong{-}Hoon Oh}, \bibinfo{person}{Ryu Iida},
  \bibinfo{person}{Canasai Kruengkrai}, {and} \bibinfo{person}{Julien
  Kloetzer}.} \bibinfo{year}{2018}\natexlab{}.
\newblock \showarticletitle{Semi-Distantly Supervised Neural Model for
  Generating Compact Answers to Open-Domain Why Questions}. In
  \bibinfo{booktitle}{\emph{AAAI}}. \bibinfo{pages}{5803--5811}.
\newblock


\bibitem[\protect\citeauthoryear{Izacard and Grave}{Izacard and Grave}{2020}]%
        {fid}
\bibfield{author}{\bibinfo{person}{Gautier Izacard} {and}
  \bibinfo{person}{Edouard Grave}.} \bibinfo{year}{2020}\natexlab{}.
\newblock \showarticletitle{Leveraging Passage Retrieval with Generative Models
  for Open Domain Question Answering}.
\newblock \bibinfo{journal}{\emph{CoRR}}  \bibinfo{volume}{abs/2007.01282}
  (\bibinfo{year}{2020}).
\newblock


\bibitem[\protect\citeauthoryear{Kingma and Welling}{Kingma and
  Welling}{2014}]%
        {iclr14-vae}
\bibfield{author}{\bibinfo{person}{Diederik~P. Kingma} {and}
  \bibinfo{person}{Max Welling}.} \bibinfo{year}{2014}\natexlab{}.
\newblock \showarticletitle{Auto-Encoding Variational Bayes}. In
  \bibinfo{booktitle}{\emph{ICLR}}.
\newblock


\bibitem[\protect\citeauthoryear{Kulkarni, Mehta, Garg, Bansal, Rasiwasia, and
  Sengamedu}{Kulkarni et~al\mbox{.}}{2019}]%
        {www19-pqa}
\bibfield{author}{\bibinfo{person}{Ashish Kulkarni}, \bibinfo{person}{Kartik
  Mehta}, \bibinfo{person}{Shweta Garg}, \bibinfo{person}{Vidit Bansal},
  \bibinfo{person}{Nikhil Rasiwasia}, {and} \bibinfo{person}{Srinivasan~H.
  Sengamedu}.} \bibinfo{year}{2019}\natexlab{}.
\newblock \showarticletitle{ProductQnA: Answering User Questions on E-Commerce
  Product Pages}. In \bibinfo{booktitle}{\emph{WWW}}.
  \bibinfo{pages}{354--360}.
\newblock


\bibitem[\protect\citeauthoryear{Lewis, Perez, Piktus, Petroni, Karpukhin,
  Goyal, K{\"{u}}ttler, Lewis, Yih, Rockt{\"{a}}schel, Riedel, and Kiela}{Lewis
  et~al\mbox{.}}{2020}]%
        {nips20-rag}
\bibfield{author}{\bibinfo{person}{Patrick S.~H. Lewis}, \bibinfo{person}{Ethan
  Perez}, \bibinfo{person}{Aleksandra Piktus}, \bibinfo{person}{Fabio Petroni},
  \bibinfo{person}{Vladimir Karpukhin}, \bibinfo{person}{Naman Goyal},
  \bibinfo{person}{Heinrich K{\"{u}}ttler}, \bibinfo{person}{Mike Lewis},
  \bibinfo{person}{Wen{-}tau Yih}, \bibinfo{person}{Tim Rockt{\"{a}}schel},
  \bibinfo{person}{Sebastian Riedel}, {and} \bibinfo{person}{Douwe Kiela}.}
  \bibinfo{year}{2020}\natexlab{}.
\newblock \showarticletitle{Retrieval-Augmented Generation for
  Knowledge-Intensive {NLP} Tasks}. In \bibinfo{booktitle}{\emph{NIPS}}.
\newblock


\bibitem[\protect\citeauthoryear{Li, Li, and Zong}{Li et~al\mbox{.}}{2019a}]%
        {aaai19-review-summ}
\bibfield{author}{\bibinfo{person}{Junjie Li}, \bibinfo{person}{Haoran Li},
  {and} \bibinfo{person}{Chengqing Zong}.} \bibinfo{year}{2019}\natexlab{a}.
\newblock \showarticletitle{Towards Personalized Review Summarization via
  User-Aware Sequence Network}. In \bibinfo{booktitle}{\emph{AAAI}}.
  \bibinfo{pages}{6690--6697}.
\newblock


\bibitem[\protect\citeauthoryear{Li, Li, Zhao, He, Wei, Yuan, and Wen}{Li
  et~al\mbox{.}}{2020}]%
        {cikm20-review-gen}
\bibfield{author}{\bibinfo{person}{Junyi Li}, \bibinfo{person}{Siqing Li},
  \bibinfo{person}{Wayne~Xin Zhao}, \bibinfo{person}{Gaole He},
  \bibinfo{person}{Zhicheng Wei}, \bibinfo{person}{Nicholas~Jing Yuan}, {and}
  \bibinfo{person}{Ji{-}Rong Wen}.} \bibinfo{year}{2020}\natexlab{}.
\newblock \showarticletitle{Knowledge-Enhanced Personalized Review Generation
  with Capsule Graph Neural Network}. In \bibinfo{booktitle}{\emph{CIKM}}.
\newblock


\bibitem[\protect\citeauthoryear{Li and Tuzhilin}{Li and Tuzhilin}{2019}]%
        {emnlp19-review-gen}
\bibfield{author}{\bibinfo{person}{Pan Li} {and} \bibinfo{person}{Alexander
  Tuzhilin}.} \bibinfo{year}{2019}\natexlab{}.
\newblock \showarticletitle{Towards Controllable and Personalized Review
  Generation}. In \bibinfo{booktitle}{\emph{EMNLP-IJCNLP}}.
  \bibinfo{pages}{3235--3243}.
\newblock


\bibitem[\protect\citeauthoryear{Li, Wang, Bing, and Lam}{Li
  et~al\mbox{.}}{2019b}]%
        {www19-tips}
\bibfield{author}{\bibinfo{person}{Piji Li}, \bibinfo{person}{Zihao Wang},
  \bibinfo{person}{Lidong Bing}, {and} \bibinfo{person}{Wai Lam}.}
  \bibinfo{year}{2019}\natexlab{b}.
\newblock \showarticletitle{Persona-Aware Tips Generation?}. In
  \bibinfo{booktitle}{\emph{WWW}}. \bibinfo{pages}{1006--1016}.
\newblock


\bibitem[\protect\citeauthoryear{Liu, Lowe, Serban, Noseworthy, Charlin, and
  Pineau}{Liu et~al\mbox{.}}{2016}]%
        {DBLP:conf/emnlp/LiuLSNCP16}
\bibfield{author}{\bibinfo{person}{Chia{-}Wei Liu}, \bibinfo{person}{Ryan
  Lowe}, \bibinfo{person}{Iulian Serban}, \bibinfo{person}{Michael Noseworthy},
  \bibinfo{person}{Laurent Charlin}, {and} \bibinfo{person}{Joelle Pineau}.}
  \bibinfo{year}{2016}\natexlab{}.
\newblock \showarticletitle{How {NOT} To Evaluate Your Dialogue System: An
  Empirical Study of Unsupervised Evaluation Metrics for Dialogue Response
  Generation}. In \bibinfo{booktitle}{\emph{EMNLP}}.
  \bibinfo{pages}{2122--2132}.
\newblock


\bibitem[\protect\citeauthoryear{Majumder, Li, Ni, and McAuley}{Majumder
  et~al\mbox{.}}{2019}]%
        {emnlp19-recipe}
\bibfield{author}{\bibinfo{person}{Bodhisattwa~Prasad Majumder},
  \bibinfo{person}{Shuyang Li}, \bibinfo{person}{Jianmo Ni}, {and}
  \bibinfo{person}{Julian~J. McAuley}.} \bibinfo{year}{2019}\natexlab{}.
\newblock \showarticletitle{Generating Personalized Recipes from Historical
  User Preferences}. In \bibinfo{booktitle}{\emph{EMNLP-IJCNLP}}.
  \bibinfo{pages}{5975--5981}.
\newblock


\bibitem[\protect\citeauthoryear{McAuley and Yang}{McAuley and Yang}{2016}]%
        {www16-amazon-qa}
\bibfield{author}{\bibinfo{person}{Julian McAuley} {and} \bibinfo{person}{Alex
  Yang}.} \bibinfo{year}{2016}\natexlab{}.
\newblock \showarticletitle{Addressing Complex and Subjective Product-Related
  Queries with Customer Reviews}. In \bibinfo{booktitle}{\emph{WWW}}.
  \bibinfo{pages}{625--635}.
\newblock


\bibitem[\protect\citeauthoryear{Miao, Grefenstette, and Blunsom}{Miao
  et~al\mbox{.}}{2017}]%
        {icml17-gsm}
\bibfield{author}{\bibinfo{person}{Yishu Miao}, \bibinfo{person}{Edward
  Grefenstette}, {and} \bibinfo{person}{Phil Blunsom}.}
  \bibinfo{year}{2017}\natexlab{}.
\newblock \showarticletitle{Discovering Discrete Latent Topics with Neural
  Variational Inference}. In \bibinfo{booktitle}{\emph{ICML}}.
  \bibinfo{pages}{2410--2419}.
\newblock


\bibitem[\protect\citeauthoryear{Musto, Rossiello, de~Gemmis, Lops, and
  Semeraro}{Musto et~al\mbox{.}}{2019}]%
        {recsys19-justification}
\bibfield{author}{\bibinfo{person}{Cataldo Musto}, \bibinfo{person}{Gaetano
  Rossiello}, \bibinfo{person}{Marco de Gemmis}, \bibinfo{person}{Pasquale
  Lops}, {and} \bibinfo{person}{Giovanni Semeraro}.}
  \bibinfo{year}{2019}\natexlab{}.
\newblock \showarticletitle{Combining text summarization and aspect-based
  sentiment analysis of users' reviews to justify recommendations}. In
  \bibinfo{booktitle}{\emph{RecSys}}. \bibinfo{pages}{383--387}.
\newblock


\bibitem[\protect\citeauthoryear{Nakatsuji and Okui}{Nakatsuji and
  Okui}{2020}]%
        {conclusion-ans}
\bibfield{author}{\bibinfo{person}{Makoto Nakatsuji} {and}
  \bibinfo{person}{Sohei Okui}.} \bibinfo{year}{2020}\natexlab{}.
\newblock \showarticletitle{Conclusion-Supplement Answer Generation for
  Non-Factoid Questions}. In \bibinfo{booktitle}{\emph{AAAI}}.
  \bibinfo{pages}{8520--8527}.
\newblock


\bibitem[\protect\citeauthoryear{Nguyen, Rosenberg, Song, Gao, Tiwary,
  Majumder, and Deng}{Nguyen et~al\mbox{.}}{2016}]%
        {msmarco}
\bibfield{author}{\bibinfo{person}{Tri Nguyen}, \bibinfo{person}{Mir
  Rosenberg}, \bibinfo{person}{Xia Song}, \bibinfo{person}{Jianfeng Gao},
  \bibinfo{person}{Saurabh Tiwary}, \bibinfo{person}{Rangan Majumder}, {and}
  \bibinfo{person}{Li Deng}.} \bibinfo{year}{2016}\natexlab{}.
\newblock \showarticletitle{{MS} {MARCO:} {A} Human Generated MAchine Reading
  COmprehension Dataset}. In \bibinfo{booktitle}{\emph{NIPS}}
  \emph{(\bibinfo{series}{{CEUR} Workshop Proceedings})},
  Vol.~\bibinfo{volume}{1773}.
\newblock


\bibitem[\protect\citeauthoryear{Ni, Li, and McAuley}{Ni et~al\mbox{.}}{2019}]%
        {emnlp19-amazon-review}
\bibfield{author}{\bibinfo{person}{Jianmo Ni}, \bibinfo{person}{Jiacheng Li},
  {and} \bibinfo{person}{Julian~J. McAuley}.} \bibinfo{year}{2019}\natexlab{}.
\newblock \showarticletitle{Justifying Recommendations using Distantly-Labeled
  Reviews and Fine-Grained Aspects}. In
  \bibinfo{booktitle}{\emph{EMNLP-IJCNLP}}. \bibinfo{pages}{188--197}.
\newblock


\bibitem[\protect\citeauthoryear{Ni and McAuley}{Ni and McAuley}{2018}]%
        {acl18-review-gen}
\bibfield{author}{\bibinfo{person}{Jianmo Ni} {and} \bibinfo{person}{Julian~J.
  McAuley}.} \bibinfo{year}{2018}\natexlab{}.
\newblock \showarticletitle{Personalized Review Generation By Expanding Phrases
  and Attending on Aspect-Aware Representations}. In
  \bibinfo{booktitle}{\emph{ACL}}. \bibinfo{pages}{706--711}.
\newblock


\bibitem[\protect\citeauthoryear{Nishida, Saito, Nishida, Shinoda, Otsuka,
  Asano, and Tomita}{Nishida et~al\mbox{.}}{2019}]%
        {acl19-answer-gen}
\bibfield{author}{\bibinfo{person}{Kyosuke Nishida}, \bibinfo{person}{Itsumi
  Saito}, \bibinfo{person}{Kosuke Nishida}, \bibinfo{person}{Kazutoshi
  Shinoda}, \bibinfo{person}{Atsushi Otsuka}, \bibinfo{person}{Hisako Asano},
  {and} \bibinfo{person}{Junji Tomita}.} \bibinfo{year}{2019}\natexlab{}.
\newblock \showarticletitle{Multi-style Generative Reading Comprehension}. In
  \bibinfo{booktitle}{\emph{ACL}}. \bibinfo{pages}{2273--2284}.
\newblock


\bibitem[\protect\citeauthoryear{Rajpurkar, Zhang, Lopyrev, and
  Liang}{Rajpurkar et~al\mbox{.}}{2016}]%
        {squad}
\bibfield{author}{\bibinfo{person}{Pranav Rajpurkar}, \bibinfo{person}{Jian
  Zhang}, \bibinfo{person}{Konstantin Lopyrev}, {and} \bibinfo{person}{Percy
  Liang}.} \bibinfo{year}{2016}\natexlab{}.
\newblock \showarticletitle{SQuAD: 100, 000+ Questions for Machine
  Comprehension of Text}. In \bibinfo{booktitle}{\emph{EMNLP}}.
  \bibinfo{pages}{2383--2392}.
\newblock


\bibitem[\protect\citeauthoryear{Sachdeva and McAuley}{Sachdeva and
  McAuley}{2020}]%
        {sigir20-review4rec}
\bibfield{author}{\bibinfo{person}{Noveen Sachdeva} {and}
  \bibinfo{person}{Julian~J. McAuley}.} \bibinfo{year}{2020}\natexlab{}.
\newblock \showarticletitle{How Useful are Reviews for Recommendation? {A}
  Critical Review and Potential Improvements}. In
  \bibinfo{booktitle}{\emph{SIGIR}}. \bibinfo{pages}{1845--1848}.
\newblock


\bibitem[\protect\citeauthoryear{See, Liu, and Manning}{See
  et~al\mbox{.}}{2017}]%
        {acl17-pointer}
\bibfield{author}{\bibinfo{person}{Abigail See}, \bibinfo{person}{Peter~J.
  Liu}, {and} \bibinfo{person}{Christopher~D. Manning}.}
  \bibinfo{year}{2017}\natexlab{}.
\newblock \showarticletitle{Get To The Point: Summarization with
  Pointer-Generator Networks}. In \bibinfo{booktitle}{\emph{ACL}}.
  \bibinfo{pages}{1073--1083}.
\newblock


\bibitem[\protect\citeauthoryear{Seo, Kembhavi, Farhadi, and Hajishirzi}{Seo
  et~al\mbox{.}}{2017}]%
        {iclr17-bidaf}
\bibfield{author}{\bibinfo{person}{Min~Joon Seo}, \bibinfo{person}{Aniruddha
  Kembhavi}, \bibinfo{person}{Ali Farhadi}, {and} \bibinfo{person}{Hannaneh
  Hajishirzi}.} \bibinfo{year}{2017}\natexlab{}.
\newblock \showarticletitle{Bidirectional Attention Flow for Machine
  Comprehension}. In \bibinfo{booktitle}{\emph{ICLR}}.
\newblock


\bibitem[\protect\citeauthoryear{Shen, Deng, Yang, Li, Du, Fan, and Lei}{Shen
  et~al\mbox{.}}{2018}]%
        {kablstm}
\bibfield{author}{\bibinfo{person}{Ying Shen}, \bibinfo{person}{Yang Deng},
  \bibinfo{person}{Min Yang}, \bibinfo{person}{Yaliang Li},
  \bibinfo{person}{Nan Du}, \bibinfo{person}{Wei Fan}, {and}
  \bibinfo{person}{Kai Lei}.} \bibinfo{year}{2018}\natexlab{}.
\newblock \showarticletitle{Knowledge-aware Attentive Neural Network for
  Ranking Question Answer Pairs}. In \bibinfo{booktitle}{\emph{SIGIR}}.
  \bibinfo{pages}{901--904}.
\newblock


\bibitem[\protect\citeauthoryear{Song, Wang, Zhang, Liu, and Liu}{Song
  et~al\mbox{.}}{2020}]%
        {acl2020-dialog}
\bibfield{author}{\bibinfo{person}{Haoyu Song}, \bibinfo{person}{Yan Wang},
  \bibinfo{person}{Weinan Zhang}, \bibinfo{person}{Xiaojiang Liu}, {and}
  \bibinfo{person}{Ting Liu}.} \bibinfo{year}{2020}\natexlab{}.
\newblock \showarticletitle{Generate, Delete and Rewrite: {A} Three-Stage
  Framework for Improving Persona Consistency of Dialogue Generation}. In
  \bibinfo{booktitle}{\emph{ACL}}. \bibinfo{pages}{5821--5831}.
\newblock


\bibitem[\protect\citeauthoryear{Song, Zhang, Cui, Wang, and Liu}{Song
  et~al\mbox{.}}{2019}]%
        {ijcai19-dialog}
\bibfield{author}{\bibinfo{person}{Haoyu Song}, \bibinfo{person}{Weinan Zhang},
  \bibinfo{person}{Yiming Cui}, \bibinfo{person}{Dong Wang}, {and}
  \bibinfo{person}{Ting Liu}.} \bibinfo{year}{2019}\natexlab{}.
\newblock \showarticletitle{Exploiting Persona Information for Diverse
  Generation of Conversational Responses}. In
  \bibinfo{booktitle}{\emph{IJCAI}}. \bibinfo{pages}{5190--5196}.
\newblock


\bibitem[\protect\citeauthoryear{Sun, Wu, Zhang, Fu, Hong, and Wang}{Sun
  et~al\mbox{.}}{2020}]%
        {www20-review-gen}
\bibfield{author}{\bibinfo{person}{Peijie Sun}, \bibinfo{person}{Le Wu},
  \bibinfo{person}{Kun Zhang}, \bibinfo{person}{Yanjie Fu},
  \bibinfo{person}{Richang Hong}, {and} \bibinfo{person}{Meng Wang}.}
  \bibinfo{year}{2020}\natexlab{}.
\newblock \showarticletitle{Dual Learning for Explainable Recommendation:
  Towards Unifying User Preference Prediction and Review Generation}. In
  \bibinfo{booktitle}{\emph{WWW}}. \bibinfo{pages}{837--847}.
\newblock


\bibitem[\protect\citeauthoryear{Vaswani, Shazeer, Parmar, Uszkoreit, Jones,
  Gomez, Kaiser, and Polosukhin}{Vaswani et~al\mbox{.}}{2017}]%
        {nips17-transformer}
\bibfield{author}{\bibinfo{person}{Ashish Vaswani}, \bibinfo{person}{Noam
  Shazeer}, \bibinfo{person}{Niki Parmar}, \bibinfo{person}{Jakob Uszkoreit},
  \bibinfo{person}{Llion Jones}, \bibinfo{person}{Aidan~N. Gomez},
  \bibinfo{person}{Lukasz Kaiser}, {and} \bibinfo{person}{Illia Polosukhin}.}
  \bibinfo{year}{2017}\natexlab{}.
\newblock \showarticletitle{Attention is All you Need}. In
  \bibinfo{booktitle}{\emph{NeurIPS}}. \bibinfo{pages}{5998--6008}.
\newblock


\bibitem[\protect\citeauthoryear{Wan and McAuley}{Wan and McAuley}{2016}]%
        {icdm16-amazon}
\bibfield{author}{\bibinfo{person}{Mengting Wan} {and}
  \bibinfo{person}{Julian~J. McAuley}.} \bibinfo{year}{2016}\natexlab{}.
\newblock \showarticletitle{Modeling Ambiguity, Subjectivity, and Diverging
  Viewpoints in Opinion Question Answering Systems}. In
  \bibinfo{booktitle}{\emph{ICDM}}. \bibinfo{pages}{489--498}.
\newblock


\bibitem[\protect\citeauthoryear{Wang, Raghavan, Cardie, and Castelli}{Wang
  et~al\mbox{.}}{2014}]%
        {coling14-cqa}
\bibfield{author}{\bibinfo{person}{Lu Wang}, \bibinfo{person}{Hema Raghavan},
  \bibinfo{person}{Claire Cardie}, {and} \bibinfo{person}{Vittorio Castelli}.}
  \bibinfo{year}{2014}\natexlab{}.
\newblock \showarticletitle{Query-Focused Opinion Summarization for
  User-Generated Content}. In \bibinfo{booktitle}{\emph{COLING}}.
  \bibinfo{pages}{1660--1669}.
\newblock


\bibitem[\protect\citeauthoryear{Wu, Li, Wang, Chen, Wong, Feng, Huang, and
  Wang}{Wu et~al\mbox{.}}{2020}]%
        {acl20-dialog-metrics}
\bibfield{author}{\bibinfo{person}{Bowen Wu}, \bibinfo{person}{Mengyuan Li},
  \bibinfo{person}{Zongsheng Wang}, \bibinfo{person}{Yifu Chen},
  \bibinfo{person}{Derek~F. Wong}, \bibinfo{person}{Qihang Feng},
  \bibinfo{person}{Junhong Huang}, {and} \bibinfo{person}{Baoxun Wang}.}
  \bibinfo{year}{2020}\natexlab{}.
\newblock \showarticletitle{Guiding Variational Response Generator to Exploit
  Persona}. In \bibinfo{booktitle}{\emph{ACL}}. \bibinfo{pages}{53--65}.
\newblock


\bibitem[\protect\citeauthoryear{Xu, Li, Yang, Ren, Ren, Chen, and Ma}{Xu
  et~al\mbox{.}}{2020}]%
        {ecai20-dialog}
\bibfield{author}{\bibinfo{person}{Minghong Xu}, \bibinfo{person}{Piji Li},
  \bibinfo{person}{Haoran Yang}, \bibinfo{person}{Pengjie Ren},
  \bibinfo{person}{Zhaochun Ren}, \bibinfo{person}{Zhumin Chen}, {and}
  \bibinfo{person}{Jun Ma}.} \bibinfo{year}{2020}\natexlab{}.
\newblock \showarticletitle{A Neural Topical Expansion Framework for
  Unstructured Persona-Oriented Dialogue Generation}. In
  \bibinfo{booktitle}{\emph{ECAI}}. \bibinfo{pages}{2244--2251}.
\newblock


\bibitem[\protect\citeauthoryear{Yu, Qiu, Jiang, Huang, Song, Chu, and Chen}{Yu
  et~al\mbox{.}}{2018}]%
        {wsdm18-answer-sel}
\bibfield{author}{\bibinfo{person}{Jianfei Yu}, \bibinfo{person}{Minghui Qiu},
  \bibinfo{person}{Jing Jiang}, \bibinfo{person}{Jun Huang},
  \bibinfo{person}{Shuangyong Song}, \bibinfo{person}{Wei Chu}, {and}
  \bibinfo{person}{Haiqing Chen}.} \bibinfo{year}{2018}\natexlab{}.
\newblock \showarticletitle{Modelling Domain Relationships for Transfer
  Learning on Retrieval-based Question Answering Systems in E-commerce}. In
  \bibinfo{booktitle}{\emph{WSDM}}. \bibinfo{pages}{682--690}.
\newblock


\bibitem[\protect\citeauthoryear{Yu, Zha, and Chua}{Yu et~al\mbox{.}}{2012}]%
        {emnlp12-answer-pred}
\bibfield{author}{\bibinfo{person}{Jianxing Yu}, \bibinfo{person}{Zheng{-}Jun
  Zha}, {and} \bibinfo{person}{Tat{-}Seng Chua}.}
  \bibinfo{year}{2012}\natexlab{}.
\newblock \showarticletitle{Answering Opinion Questions on Products by
  Exploiting Hierarchical Organization of Consumer Reviews}. In
  \bibinfo{booktitle}{\emph{EMNLP-CoNLL}}. \bibinfo{pages}{391--401}.
\newblock


\bibitem[\protect\citeauthoryear{Yu and Lam}{Yu and Lam}{2018}]%
        {wsdm18-answer-pred}
\bibfield{author}{\bibinfo{person}{Qian Yu} {and} \bibinfo{person}{Wai Lam}.}
  \bibinfo{year}{2018}\natexlab{}.
\newblock \showarticletitle{Review-Aware Answer Prediction for Product-Related
  Questions Incorporating Aspects}. In \bibinfo{booktitle}{\emph{WSDM}}.
  \bibinfo{pages}{691--699}.
\newblock


\bibitem[\protect\citeauthoryear{Zeng, Xu, and Ai}{Zeng et~al\mbox{.}}{2021}]%
        {ecir21-review4rec}
\bibfield{author}{\bibinfo{person}{Hansi Zeng}, \bibinfo{person}{Zhichao Xu},
  {and} \bibinfo{person}{Qingyao Ai}.} \bibinfo{year}{2021}\natexlab{}.
\newblock \showarticletitle{A Zero Attentive Relevance Matching Network for
  Review Modeling in Recommendation System}. In
  \bibinfo{booktitle}{\emph{ECIR}}. \bibinfo{pages}{724--739}.
\newblock


\bibitem[\protect\citeauthoryear{Zeng, Abuduweili, Li, and Yang}{Zeng
  et~al\mbox{.}}{2019}]%
        {acl19-comment}
\bibfield{author}{\bibinfo{person}{Wenhuan Zeng}, \bibinfo{person}{Abulikemu
  Abuduweili}, \bibinfo{person}{Lei Li}, {and} \bibinfo{person}{Pengcheng
  Yang}.} \bibinfo{year}{2019}\natexlab{}.
\newblock \showarticletitle{Automatic Generation of Personalized Comment Based
  on User Profile}. In \bibinfo{booktitle}{\emph{{ACL} 2019, Volume 2: Student
  Research Workshop}}. \bibinfo{pages}{229--235}.
\newblock


\bibitem[\protect\citeauthoryear{Zhang, Dinan, Urbanek, Szlam, Kiela, and
  Weston}{Zhang et~al\mbox{.}}{2018}]%
        {acl18-dialog}
\bibfield{author}{\bibinfo{person}{Saizheng Zhang}, \bibinfo{person}{Emily
  Dinan}, \bibinfo{person}{Jack Urbanek}, \bibinfo{person}{Arthur Szlam},
  \bibinfo{person}{Douwe Kiela}, {and} \bibinfo{person}{Jason Weston}.}
  \bibinfo{year}{2018}\natexlab{}.
\newblock \showarticletitle{Personalizing Dialogue Agents: {I} have a dog, do
  you have pets too?}. In \bibinfo{booktitle}{\emph{ACL}}.
  \bibinfo{pages}{2204--2213}.
\newblock


\bibitem[\protect\citeauthoryear{Zhang, Yao, Sun, and Tay}{Zhang
  et~al\mbox{.}}{2019}]%
        {dl-rec}
\bibfield{author}{\bibinfo{person}{Shuai Zhang}, \bibinfo{person}{Lina Yao},
  \bibinfo{person}{Aixin Sun}, {and} \bibinfo{person}{Yi Tay}.}
  \bibinfo{year}{2019}\natexlab{}.
\newblock \showarticletitle{Deep Learning Based Recommender System: {A} Survey
  and New Perspectives}.
\newblock \bibinfo{journal}{\emph{{ACM} Comput. Surv.}} \bibinfo{volume}{52},
  \bibinfo{number}{1} (\bibinfo{year}{2019}), \bibinfo{pages}{5:1--5:38}.
\newblock


\bibitem[\protect\citeauthoryear{Zhang, Deng, and Lam}{Zhang
  et~al\mbox{.}}{2020a}]%
        {sigir20-answer-sel}
\bibfield{author}{\bibinfo{person}{Wenxuan Zhang}, \bibinfo{person}{Yang Deng},
  {and} \bibinfo{person}{Wai Lam}.} \bibinfo{year}{2020}\natexlab{a}.
\newblock \showarticletitle{Answer Ranking for Product-Related Questions via
  Multiple Semantic Relations Modeling}. In \bibinfo{booktitle}{\emph{SIGIR}}.
  \bibinfo{pages}{569--578}.
\newblock


\bibitem[\protect\citeauthoryear{Zhang, Deng, Ma, and Lam}{Zhang
  et~al\mbox{.}}{2020b}]%
        {answerfact}
\bibfield{author}{\bibinfo{person}{Wenxuan Zhang}, \bibinfo{person}{Yang Deng},
  \bibinfo{person}{Jing Ma}, {and} \bibinfo{person}{Wai Lam}.}
  \bibinfo{year}{2020}\natexlab{b}.
\newblock \showarticletitle{AnswerFact: Fact Checking in Product Question
  Answering}. In \bibinfo{booktitle}{\emph{EMNLP}}.
  \bibinfo{pages}{2407--2417}.
\newblock


\bibitem[\protect\citeauthoryear{Zhang, Lam, Deng, and Ma}{Zhang
  et~al\mbox{.}}{2020c}]%
        {www2020-answer-help}
\bibfield{author}{\bibinfo{person}{Wenxuan Zhang}, \bibinfo{person}{Wai Lam},
  \bibinfo{person}{Yang Deng}, {and} \bibinfo{person}{Jing Ma}.}
  \bibinfo{year}{2020}\natexlab{c}.
\newblock \showarticletitle{Review-guided Helpful Answer Identification in
  E-commerce}. In \bibinfo{booktitle}{\emph{WWW}}. \bibinfo{pages}{2620--2626}.
\newblock


\bibitem[\protect\citeauthoryear{Zhao, Fu, Song, Sakai, Chen, Xie, and
  Qian}{Zhao et~al\mbox{.}}{2019a}]%
        {tist-reason}
\bibfield{author}{\bibinfo{person}{Guoshuai Zhao}, \bibinfo{person}{Hao Fu},
  \bibinfo{person}{Ruihua Song}, \bibinfo{person}{Tetsuya Sakai},
  \bibinfo{person}{Zhongxia Chen}, \bibinfo{person}{Xing Xie}, {and}
  \bibinfo{person}{Xueming Qian}.} \bibinfo{year}{2019}\natexlab{a}.
\newblock \showarticletitle{Personalized Reason Generation for Explainable Song
  Recommendation}.
\newblock \bibinfo{journal}{\emph{{ACM} Trans. Intell. Syst. Technol.}}
  \bibinfo{volume}{10}, \bibinfo{number}{4} (\bibinfo{year}{2019}),
  \bibinfo{pages}{41:1--41:21}.
\newblock


\bibitem[\protect\citeauthoryear{Zhao, Guan, and Sun}{Zhao
  et~al\mbox{.}}{2019b}]%
        {kdd19-answer-sel}
\bibfield{author}{\bibinfo{person}{Jie Zhao}, \bibinfo{person}{Ziyu Guan},
  {and} \bibinfo{person}{Huan Sun}.} \bibinfo{year}{2019}\natexlab{b}.
\newblock \showarticletitle{Riker: Mining Rich Keyword Representations for
  Interpretable Product Question Answering}. In
  \bibinfo{booktitle}{\emph{KDD}}. \bibinfo{pages}{1389--1398}.
\newblock


\end{thebibliography}

\end{document}